\documentclass{jair}

\setcopyright{cc}
\acmDOI{10.1613/jair.1.xxxxx}

\JAIRAE{Roberta Calegari}
\JAIRTrack{AI and Society Track} 
\acmVolume{4}
\acmArticle{6}
\acmMonth{8}
\acmYear{2026}

\RequirePackage[
  datamodel=acmdatamodel,
  style=acmauthoryear,
  backend=biber,
  giveninits=true,
  uniquename=init
  ]{biblatex}

\usepackage[utf8]{inputenc} 
\usepackage[T1]{fontenc}    
\usepackage{hyperref}       
\usepackage{url}            
\usepackage{booktabs}       
\usepackage{amsfonts}       
\usepackage{nicefrac}       
\usepackage{microtype}      
\usepackage{amsmath}
\usepackage{graphicx}
\usepackage{xcolor}
\usepackage{colortbl}
\usepackage{multirow}
\usepackage{float}
\usepackage{subcaption}

\usepackage{enumitem}
\setlist[itemize]{leftmargin=*}
\setlist[enumerate]{leftmargin=*}
\usepackage{array} 
\newcolumntype{P}[1]{>{\centering\arraybackslash}p{#1}}
\usepackage{pifont}

\newcommand{\bi}{\begin{itemize}}
\newcommand{\ei}{\end{itemize}}
\newcommand{\be}{\begin{enumerate}}
\newcommand{\ee}{\end{enumerate}}

\usepackage{makecell}
\usepackage[linesnumbered,ruled,vlined]{algorithm2e}
\usepackage{algorithmic}
\usepackage[skins]{tcolorbox}

\usepackage{wrapfig}

\usepackage[skins]{tcolorbox}

\usepackage{diagbox}

\begin{document}

\title[Differential Parity]{Differential Parity: Relative Fairness Between Two Sets of Decisions}


\author{Zhe Yu}
\authornote{Corresponding Author.}
\orcid{0000-0002-6841-1725}
\email{zxyvse@rit.edu}
\author{Xiaoyin Xi}
\orcid{0009-0001-8313-309X}
\email{xx4455@rit.edu}
\author{Pranam Prakash Shetty}
\orcid{0009-0008-2809-8499}
\email{ps9960@rit.edu}
\affiliation{%
  \institution{Rochester Institute of Technology}
  \city{Rochester}
  \state{New York}
  \country{USA}
}

\renewcommand{\shortauthors}{Yu, Xi \& Shetty}

\begin{abstract}
{\bf Background:} 
    With AI systems increasingly being applied to assist humans in decision-making processes such as talent hiring, school admissions, and loan approvals, there is a growing need to ensure that the resulting decisions are fair. A major challenge in analyzing fairness is that standards are highly subjective and context-dependent —- there is no consensus on what absolute fairness means in every scenario. Moreover, different standards of fairness often conflict with each other.
    
    {\bf Objectives:}
    To address this issue, this work aims to evaluate the relative fairness between decisions.
    
    {\bf Methods:}
    Instead of defining what constitutes ``absolutely'' fair decisions, we propose assessing the relative fairness of one decision set against another using differential parity —- two sets of decisions are considered relatively fair with respect to each other if and only if the difference between them is independent of a given sensitive attribute. The proposed notion of differential parity fairness offers three key benefits: (1) it avoids the ambiguity and contradictions inherent in defining ``absolutely'' fair decisions; (2) it reveals relative preferences and biases between two decision sets; and (3) it can serve as a new notion of group fairness when a reference set of decisions (e.g., ground truth) is available. One limitation of differential parity is that the two sets of decisions being compared must be made on the same data subjects. To overcome this limitation, we propose to utilize a machine learning model to bridge the gap between the two sets of decisions made on different data and approximate the differential parity metrics. In addition to differential parity and inspired by the statistical parity fairness notion, we also define relative statistical parity -- the difference between the means of two sets of decisions is required to be independent of the sensitive attribute -- as a weaker notion of relative fairness compared to differential parity. 
    
    {\bf Results:}
    Theoretically, we show how the proposed metrics statistically evaluate differential parity and relative statistical parity. We also proved the feasibility of using the proposed biased bridge algorithm to approximate differential parity metrics between decisions made on different data. Empirically, we evaluated the Type I and Type II error rates of differential parity and relative statistical parity both between decisions made on the same data and on different data. Experimental results suggest that differential parity outperforms relative statistical parity by having a much lower Type II error rate in both scenarios.
    
    {\bf Conclusions:}
    With lower than $0.1$ Type I and Type II error rates in both scenarios, the effectiveness of differential parity demonstrated in this article suggests that it is feasible and beneficial to evaluate relative bias between decisions made by different entities. We expect this to pave the way for the analysis of relative fairness in AI and beyond.
\end{abstract}
 
\received{08 December 2025}
\received[accepted]{14 April 2026}
\maketitle

\section{Introduction}\label{sec:introduction}

Recently, much research has focused on mitigating bias and discrimination in AI systems. This is because AI systems are increasingly being used to make decisions that affect people's lives and sometimes the learned models behave in a biased manner that gives undue advantages to a specific group of people (where those groups are determined by sex, race, etc.). Such biased decisions can have serious consequences in domains such as health care resource allocation~\cite{obermeyer2019dissecting}, consumer lending~\cite{bartlett2022consumer}, criminal justice risk assessment~\cite{chouldechova2017fair}, and hiring~\cite{raghavan2020mitigating}.

Although much research effort has focused on detecting and mitigating algorithmic bias in machine learning models, there are processes where the human makes the final decision. For example, a human resource professional may make the final interview decision while algorithmic tools are used to rank applicants or assess them in the hiring pipeline~\cite{raghavan2020mitigating}. In these scenarios, reducing bias in machine learning models alone does not make these processes fair as long as the responsible human continues to make biased final decisions. In addition, the machine learning model trained on biased human decisions will also inherit the bias~\cite{sharma2020data}. Such inherited biases may not be detected by algorithmic bias metrics such as equalized odds since the ground truth labels used to evaluate the algorithmic bias are also biased human decisions. To address this problem, this paper also evaluates biases in human decisions beyond algorithmic bias.

One major challenge for analyzing fairness in decisions is that the standards for ``absolute'' fairness are highly subjective and contextual~\cite{abu2020contextual}. For example, \textbf{College 1} demands equality and admits students without considering gender/race (fairness through unawareness~\cite{dwork2012fairness}) while \textbf{College 2} demands equity and admits the same percentage of students from each gender/race group (statistical parity in group fairness~\cite{dwork2012fairness}). Neither \textbf{College 1} or \textbf{2} is wrong, but these are often contradictory standards and lead to very different outcomes -- they are impossible to satisfy simultaneously~\cite{friedler2021possibility}. 

One way to face this challenge is to test relative fairness between different decision sets. That is, instead of defining what are ``absolutely'' fair decisions, we measure the relative fairness of one decision set against another. In the field of machine learning algorithmic fairness~\cite{hardt2016equality}, two group fairness criteria were proposed to measure whether a machine learning model's prediction $R$ is relatively fair compared to a set of reference set $Y$ (ground truth) over a certain sensitive attribute $A$. These are (1) separation $R\perp A \mid Y$ that requires the model prediction $R$ to be conditionally independent of the sensitive attribute $A$ given the reference set $Y$; and (2) sufficiency $Y\perp A \mid R$ that requires the reference set $Y$ to be conditionally independent of the sensitive attribute $A$ given the model prediction $R$. In the case of binary classification, group fairness notion equalized odds (or true positive rate parity and false positive rate parity) ensures separation while predictive parity ensures sufficiency~\cite{hardt2016equality}. However, separation and sufficiency cannot be satisfied simultaneously when $A$ and $Y$ are not statistically independent (unless $R=Y$ is a perfect predictor)~\cite{friedler2021possibility}. In addition, it is not always possible to have a ground truth reference set $Y$ of ``absolutely'' fair decisions. In the comparison of two decision sets, the choice of which one serves as the reference set $Y$ largely affects the result of separation and sufficiency. For example, the results can be that Decision Set A satisfies separation but not sufficiency when compared to a reference Decision Set B while Decision Set B satisfies sufficiency but not separation when compared to the reference Decision Set A. It is hard to tell whether these two decision sets are relatively fair to each other.

In this work, we propose a novel and general relative fairness notion called differential parity. Differential parity requires the difference between two sets of decisions to be independent of a certain sensitive attribute $(R_0 - R_1) \perp A$. Different from separation and sufficiency, which decision set is chosen as $R_0$ or $R_1$ does not affect whether differential parity is satisfied. In the case that differential parity is violated, its result can also tell the relative preference over the sensitive attribute $A$ between $R_0$ and $R_1$ -- one decision set is constantly overrating one sensitive group over the other when compared to the other decision set. For example, \textbf{Human Resource (HR) 0} always rates female candidates one point higher than \textbf{HR 1}'s ratings and \textbf{HR 0} also rates male candidates two points lower than \textbf{HR 1}'s ratings. In this case, there is a differential parity between \textbf{HR 0} and \textbf{HR 1}'s ratings over gender and the sign of its value tells that \textbf{HR 0} is overrating female candidates over male candidates compared to \textbf{HR 1} (or \textbf{HR 1} is overrating male candidates over female candidates compared to \textbf{HR 0}). Although we do not know whether \textbf{HR 0} or \textbf{HR 1} is fairer, we do know the relative bias between \textbf{HR 0} and \textbf{1} -- \textbf{HR 0} prefers female candidates more than \textbf{HR 1}. This definition of differential parity has the following benefits: (1) it avoids the ambiguous and contradictory definition of what absolutely fair decisions are; (2) when a reference set (of ground truth or reliable fair decisions) is available, differential parity can serve as a new group fairness notion (similar to but different from separation and sufficiency); (3) even when no reference set is available, it reveals the relative preference or bias between different decision sets. For the previous example, \textbf{College 1} can use a reference set where admission decisions were made regardless of gender/race and test its admission decisions against it with differential parity while \textbf{College 2} can use a different reference set where the same ratios of students were admitted in each gender/race group. 

One limitation of differential parity is that it requires the two sets of decisions to be made on the same data to calculate the difference between decisions. This will not be a problem when one of the decision sets comes from a machine learning model. However, when both decision sets are from human beings, such overlapping decisions are not always available even for the same task -- e.g. there might be multiple HRs screening for the same job application, but one application is only screened by one HR to avoid wasting of human effort. To overcome this limitation, we propose an estimation algorithm of differential parity between\textbf{ Human 0}'s decisions $R_0(x\in X_0)$ on data $X_0$ and \textbf{Human 1}'s decisions $R_1(x\in X_1)$ on data $X_1$ by (1) fitting a machine learning model $f$ with $(X_0, R_0(x\in X_0))$; (2) calculating the difference $f(x)-R_0(x)$ on $\{x\in X_0\}$ and the difference $f(x)-R_1(x)$ on $\{x\in X_1\}$. In this way, the machine learning model $f$ serves as a bridge connecting the decisions made on two different sets of data. Both theoretically and empirically, we show that, the differential parity between $R_0$ and $R_1$ can be effectively estimated with the proposed algorithm. Note that, this work does not evaluate differential parity between decisions made for different tasks or in different contexts. When we say decisions made on different data, we mean different data from the same dataset (for the same decision task). 

In addition to differential parity and inspired by the statistical parity fairness notion, we also define relative statistical parity -- the difference between the means of two sets of decisions is required to be independent of the sensitive attribute -- as a weaker notion of relative fairness compared to differential parity. The advantage of relative statistical parity is that, it can be directly evaluated between decisions made on different data. However, just like the unpaired t-tests having a lower statistical power than the paired t-tests, relative statistical parity also has a lower statistical power (or higher Type II error rate) than differential parity -- it is more likely for relative statistical parity to miss the detection of an existing relative bias when differential parity would not.

\subsection{Example Application Scenarios}
\label{sec:example}
Table~\ref{tab:scenario} demonstrates the application scenarios for separation, differential parity, and relative statistical parity.

\begin{table*}[!tbh]
\caption{Application scenarios of different relative fairness criteria.}
\small
\centering
\setlength\tabcolsep{2pt}
\label{tab:scenario}
\begin{tabular}{l|P{3.5cm}|P{3.5cm}|P{1.4cm}|P{2.3cm}|P{2.8cm}|}
& \multicolumn{2}{c|}{Scenario Description} & \multicolumn{3}{c|}{Relative Fairness} \\\cline{2-6}
           & $R_0$                       & $R_1$                       & Separation $R_0\perp A | R_1$ & Differential Parity $(R_0-R_1) \perp A$ & Relative Statistical Parity $(\overline{R}_0-\overline{R}_1) \perp A$ \\\hline
Scenario 1 & Model trained on $X_0$      & Ground Truth on $X_1$       & \checkmark                             & \checkmark          & \checkmark                                   \\
Scenario 2 & Human 0's decision on $X_0$ & Ground Truth on $X_0$       & \checkmark                             & \checkmark            & \checkmark                                 \\
Scenario 3 & Human 0's decision on $X_0$ & Ground Truth on $X_1$       & \ding{55}                             & $\sim$ & \checkmark      \\
Scenario 4 & Human 0's decision on $X_0$ & Human 1's decision on $X_0$ & \ding{55}                           & \checkmark                        & \checkmark                     \\
Scenario 5 & Human 0's decision on $X_0$ & Human 1's decision on $X_1$ & \ding{55}                              & $\sim$    & \checkmark        \\\hline
\end{tabular}
\end{table*}

\textbf{Scenario 1.} Differential parity and relative statistical parity can be applied in the same way as other relative fairness notions such as equalized odds, equal opportunity, separation, and sufficiency. This is described as Scenario 1 of Table~\ref{tab:scenario}. When a set of ground truth labels is provided (such as whether a default payment actually happened in the UCI Default of Credit Card Clients dataset~\cite{default_of_credit_card_clients_350} or whether a defendant reoffended in two years in the Compas dataset~\cite{chouldechova2017fair}, differential parity or relative statistical parity can be applied to evaluate whether a machine learning model is relatively fair with respect to the ground truth. 

\textbf{Scenario 2 and 3.} When a set of ground truth labels (of the past) is provided, differential parity and relative statistical parity can be applied to evaluate whether a human makes relatively fair decisions (on current subjects) with respect to the ground truth. A unique advantage of differential parity and relative statistical parity is that they can be evaluated when the human decisions and ground truth labels are on different data (Scenario 3). For differential parity, this is made possible by using the biased bridge algorithm and because of the subtraction property of equality. Other notions of relative fairness, such as separation, cannot evaluate two sets of decisions made on different data subjects.


\textbf{Scenario 4 and 5: Ground truth labels are not available.} This usually happens when comparing two sets of human decisions. Separation is not applicable for these scenarios since there is no way to know which human is more correct or fair (which set should be chosen as the reference set $R_1$). However, differential parity and relative statistical parity can still be applied to analyze the relative preference between the two humans, e.g. compared to \textbf{HR 1}, \textbf{HR 0} prefers women more than men. In these scenarios, a human-provided reference set is used instead of the ground truth labels. It is usually easier for humans to provide a set of acceptable decisions than to articulate the standard of what is considered as 100\% fair. For example, college admissions often rely on holistic review, in which committees weigh academic, contextual, and nonacademic factors rather than applying a single fixed rule~\cite{hossler2019study}. As an admission committee member, it is almost impossible to come up with a set of clear and executable rules/standards of what students should be admitted considering the above mentioned a variety of factors. In this case, it would be much easier and executable to spend effort and build a reference set which most decision-makers and applicants agree to be fair. Later on, this reference set (e.g. from 2022) can be used to decide whether the admission decisions in the current year (of 2023) are relatively fair in terms of differential parity (with biased bridge estimation) or relative statistical parity.

\subsection{Contributions}\label{contributions}

The contributions of this work include:
\begin{itemize}
\item
The proposed relative fairness notion differential parity covers not only algorithmic fairness but also relative fairness in human decisions.
\item
Two machine learning-based algorithms are proposed to estimate the violation of differential parity between decisions made on different data.
\item
Relative statistical parity is also proposed as a weaker version of differential parity. It can be evaluated between decisions made on different data naturally.
\item
The empirical results demonstrate the consistency and robustness of the proposed relative fairness metrics and the effectiveness of evaluating differential parity between decisions made on different data with the proposed bridging algorithms.
\item
This work provides a new way to view fairness relatively and to evaluate relative fairness.
\item
The code and data used in this paper are publicly available on GitHub at \url{https://github.com/hil-se/RelativeFairnessTesting}.
\end{itemize}

\subsection{Paper Structure}\label{structure}
The remainder of this paper is structured as follows. Section~\ref{sec:related} presents the background and related work, followed by details of the definition of differential parity and metrics to evaluate differential parity in Section~\ref{sect:methodology}. The proposed algorithms to approximate differential parity metrics between decisions made on different data are also introduced in Section~\ref{sect:methodology}. Next, the definition and metrics of relative statistical parity are described in Section~\ref{sect:methodology2}. To empirically demonstrate the effectiveness of differential parity and relative statistical parity, Section~\ref{sec:experiment} introduces two real world case studies to explore three research questions. Section~\ref{sec:results} presents the experimental results of the two case studies. Section~\ref{sec:discussion} discusses the limitations and threats to validity of this work, while Section~\ref{sec:Conclusions} summarizes the work and presents potential future work to advance it.

\section{Related Work}\label{sec:related}

\subsection{Evaluation of Fairness on One Decision Set}

Research on decision fairness is difficult. This is because the standards for ``absolute'' fairness are highly subjective and contextual~\cite{abu2020contextual}. Fairness criteria for one decision set (such as equity and equality) are often contradictory and lead to very different outcomes. Group fairness notion statistical parity~\cite{dwork2012fairness} ensures equity by requiring the outcomes to be similar across different sensitive groups while individual-level fairness notion fairness through unawareness~\cite{dwork2012fairness} ensures equality by not considering the sensitive attributes in decision-making processes. They are impossible to be satisfied simultaneously~\cite{friedler2021possibility}.

As an example, Sap et al.~\cite{sap2019risk} studied several datasets of social media posts annotated for the presence of hate speech. They showed that when the posts are written in the African American English (AAE) dialect or are authored by users self-identified as Black the posts are more likely to be labeled (both by human annotators and the models learned from them) as hate speech than if they are not written in AAE or are authored by users who self-identified as White. Two different reasons could lead to this finding -- (1) the human annotators are biased towards AAE dialect or black post authors; or (2) posts authored by users self-identified as Black or written in the AAE dialect tend to be more offensive. If it is caused by the first reason, we want to fix the bias caused by the annotators. If it is caused by the second reason, the annotations should be considered correct. However, there is no way to know for which reason this finding is made without the truth of the origin.

\subsection{Relative Fairness in Algorithmic Fairness}

Given the difficulty of obtaining the standard for ``absolute'' fairness, we turn to the testing of relative fairness between different decision sets. Previous research on machine learning algorithmic fairness has utilized a reference set to define group fairness criteria of separation and sufficiency~\cite{hardt2016equality}. Fairness notions such as equalized odds and predictive parity~\cite{hardt2016equality} are developed to evaluate separation and sufficiency, respectively, on binary sensitive attributes. The choice of reference set is crucial to such group fairness criteria since it should be considered as ``absolutely'' fair. Such ``absolutely'' fair reference set is sometimes available when testing the algorithmic fairness of a machine learning model on a dataset of ground truth. However, it is not always available especially on datasets of human decisions. Attempts have been made to estimate the ground truth by acquiring multiple annotations on the same data. Several studies have shown that, for binary labeling, 3-10 annotators per item is sufficient to obtain reliable labels (evaluated using inter-annotator agreement scores~\cite{artstein2017inter} such as Cohen's kappa and Krippendorff's alpha). \cite{dawid1979} used the EM algorithm to iteratively estimate the ground ground labels, along with the error rate of each annotator, for binary labeling problems. This model has been later extended by other researchers for other scenarios~\cite{kairam2016parting,raykar2010learning,weld2011human,ipeirotis2010quality,pasternack2010knowing,felt2014momresp,hovy-etal-2013-learning}. Liu and others have taken the more radical approach of treating the ground truth of each label not as a single value, but as a distribution over the answers that a population of annotators would provide, where the actual labels obtained are merely an observed sample of this hidden population's responses \cite{Liu2019HCOMP,Weerasooriya2020,modeling_annotators}. However, (1) high inter-annotator agreement scores do not necessarily suggest the consensus decisions are fair; and (2) these approaches are expensive and do not scale up.

\subsection{Relative Fairness Between Human Decisions}

Previously, there was no clear definition of relative fairness in human decisions. And these previous studies analyzed the data without strict controls. For example, \cite{price2010racial} concluded that white NBA referees tend to award more extra fouls to black players than black NBA referees by regressing the number of fouls called per 48 minutes for each player-game observation in which the referee participated, against an indicator variable for whether the offending player is black. The data for each referee comes from different games played by different players. Therefore, the conclusion can be misled by coincidences such as black players happened to commit more fouls in games with white referees. Another study by \cite{welch1988black} analyzed the correlation between races of judges and the punishment decisions. It concluded that black and white judges weighted case and offender information in similar ways when making punishment decisions, although black judges were more likely to sentence both black and white offenders to prison. This conclusion is also not reliable since the punishment decisions are made on different offenders.

To avoid such problem, we define a novel relative fairness criterion differential parity -- the difference between the two sets of decisions made on the same data is required to be independent of the sensitive attribute. This relative fairness definition has no assumption on the task and thus is more general than the existing ``absolute'' fairness definitions. It can be utilized to evaluate the algorithmic fairness of a machine learning model with respect to a reference set similarly as separation and sufficiency. It can also be used to evaluate the relative fairness or bias between two sets of human decisions without a ``absolutely'' fair reference. In addition to differential parity and inspired by the statistical parity fairness notion, we also define relative statistical parity -- the difference between the means of two sets of decisions is required to be independent of the sensitive attribute -- as a weaker notion of relative fairness compared to differential parity.

\subsection{Statistical Tools}
We will apply two statistical tools to evaluate the probability and magnitude of the violation of differential parity in the next Section. Here, we briefly describe the two tools: null hypothesis testing~\cite{frick1996appropriate} and effect size testing~\cite{chow1988significance}.

\textbf{Null hypothesis testing: }The null hypothesis and the alternative hypothesis are types of conjectures used in statistical tests to make statistical inferences, which are formal methods of reaching conclusions and separating scientific claims from statistical noise. The statement being tested in a test of statistical significance is called the null hypothesis. The test of significance is designed to assess the strength of the evidence against the null hypothesis~\cite{frick1996appropriate}. In our case, we construct a null hypothesis that two means are the same and test the probability $p$ of achieving the null hypothesis. If $p<0.05$, we reject the null hypothesis and confirm a difference in the two means, otherwise we accept the null hypothesis -- there is not enough evidence (either because there is no difference or because the sample size is not large enough) to support a difference between the two means. Because there is no evidence that the variances under test are the same, we applied the Welch's t-test~\cite{welch1947generalization} when the sample size is smaller than $30$ or the z-test when the sample size is larger than $30$.

\textbf{Effect size testing: }Null hypothesis testing is often critiqued for its ignoring of the effect size~\cite{cohen1994earth}. Almost any null hypothesis will get rejected given large enough sample size. As a result, we also measure the effect size of the difference of the two means, which is a quantitative measure of the magnitude of the difference regardless of the sample size~\cite{chow1988significance}. More specifically, Cohen's d test~\cite{cohen2013statistical} is applied to measure the effect size since the two means under test follow normal distributions based on the central limit theorem.

\section{Methodology: Differential Parity}
\label{sect:methodology}

In this section, we first define the notion of differential parity in Section~\ref{sect:relative fairness}, then we present the two metrics evaluating the violation of differential parity in Section~\ref{sect:metrics}. In Section~\ref{sect:framework}, two algorithms are presented to estimate the differential parity metrics between two sets of decisions made on different data.

\subsection{Relative Fairness -- Differential Parity}
\label{sect:relative fairness}
\begin{figure}[!tbh]
  \centering
  \includegraphics[width=\linewidth]{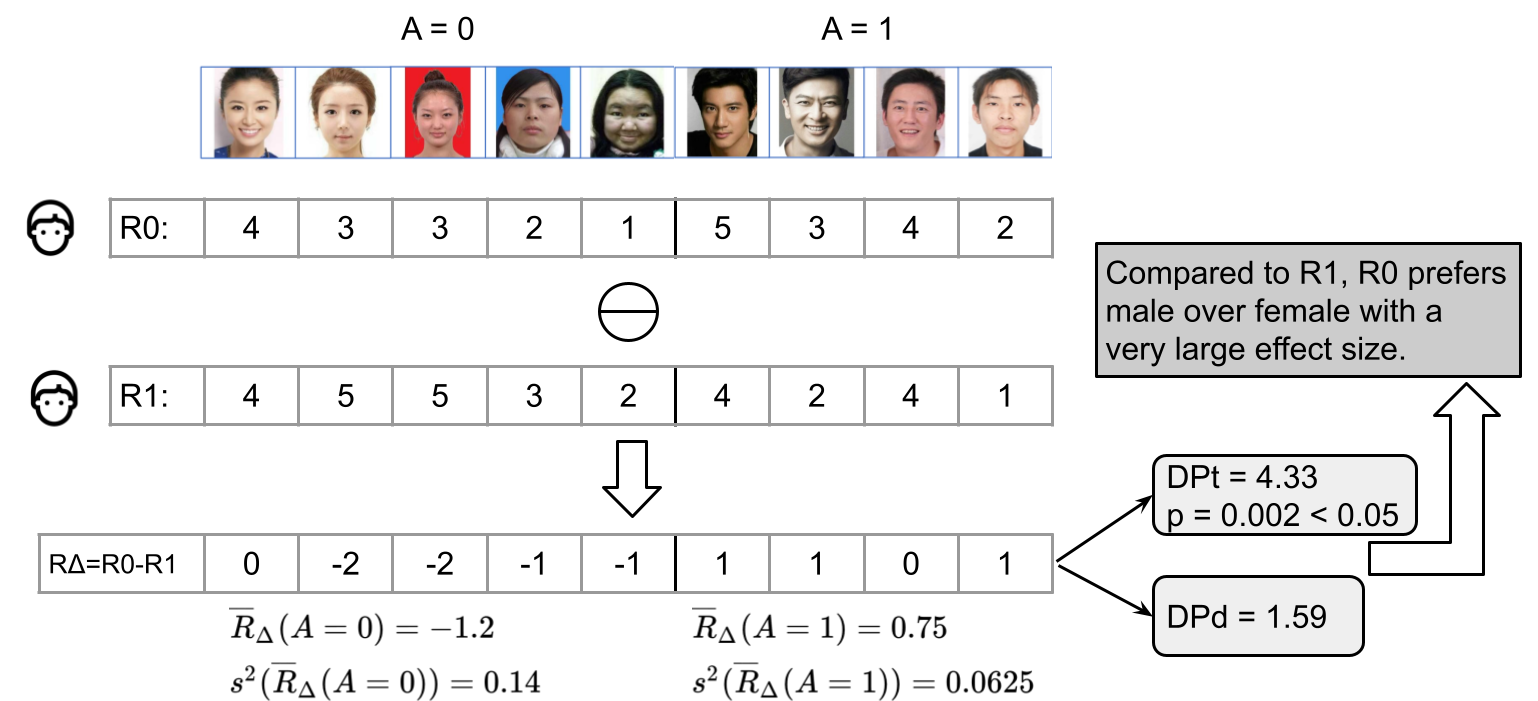}
  \caption{Illustration differential parity between two raters on face beauty.}
  \label{fig:example}
\end{figure}

Differential parity requires the pointwise difference between two decision sets are independent of a given sensitive attribute. As illustrated in Figure~\ref{fig:example}, differential parity is violated when Rater 0 rated male ($A=1$) higher and female ($A=1$) lower than Rater 1.
\begin{definition}\label{dp}
\textbf{Differential Parity.} Given a set of data $X\in\mathbb{R}^d$ with a sensitive attribute $A(x\in X)\in\mathbb{R}$, two sets of decisions $R_0(x\in X),\, R_1(x\in X) \in\mathbb{R}$ made on this data set satisfy differential parity over the sensitive attribute $A(x)$ if and only if
$$(R_0(x)-R_1(x))\perp A(x).$$
\end{definition}

\textbf{Definition~\ref{dp}} provides a general definition of differential parity that can be applied to almost any scenario. However, it is difficult to evaluate directly. Therefore, we also consider the following definition of differential parity for binary sensitive attributes.

\begin{definition}\label{dp_binary}
\textbf{Differential Parity for Binary Sensitive Attributes.} Given a set of data $X\in\mathbb{R}^d$ with a binary sensitive attribute $A(x\in X)\in\{0, 1\}$. Two sets of decisions $R_{0}(x\in X),\,R_{1}(x\in X) \in \mathbb{R}$ on this data set satisfy differential parity over the sensitive attribute $A(x)$ if and only if the difference of the decisions on each sensitive group follow the same distribution:
$$R_{\Delta}(A=0) \overset{d}{=} R_{\Delta}(A=1)$$
where
$$R_{\Delta}(A=a) = \{R_{0}(x)-R_{1}(x) \mid A(x)=a,\, x\in X\}.$$
\end{definition}
\textbf{Definition~\ref{dp_binary}} is equivalent to \textbf{Definition~\ref{dp}} in the case of binary sensitive attributes. 

\begin{theorem}\label{theorem1}
Under the assumption that the decision differences $R_{\Delta}(A=a)$ are independent and identically distributed (i.i.d.), the sampled mean of $R_{\Delta}(A=a)$ follows a normal distribution in large samples:
\begin{equation}
\label{means}
\begin{aligned}
\frac{\overline{R}_{\Delta}(A=a) - \mu(R_{\Delta}(A=a))}{\sigma(R_{\Delta}(A=a))/\sqrt{|A=a|}} \overset{d}{\rightarrow} \mathcal{N}(0, 1).
\end{aligned}
\end{equation}
where $\mu(R_{\Delta}(A=a))$ and $\sigma(R_{\Delta}(A=a))$ are the mean and standard deviation of the distribution of $R_{\Delta}(A=a)$. $|A=a|$ represents the number of samples with $A=a$.
\end{theorem}
Theorem~\ref{theorem1} can be easily proved with the central limit theorem and the law of large numbers. The assumption is also easily satisfied since the two sets of decisions are independent given any data point $x\in X$.

\subsection{Differential Parity Metrics}
\label{sect:metrics}

Based on Theorem~\ref{theorem1}, we define two metrics to measure the violation of differential parity -- whether the decision differences on each sensitive group follow the same distribution. We first measure the probability of the difference between $\overline{R}_{\Delta}(A=0)$ and $\overline{R}_{\Delta}(A=1)$ arises from random chance with null hypothesis testing~\cite{frick1996appropriate}, i.e. is there enough evidence showing that there is a difference between $\overline{R}_{\Delta}(A=0)$ and $\overline{R}_{\Delta}(A=1)$? Then we measure the strength of the difference with effect size testing~\cite{chow1988significance}, i.e. is the difference large enough to be considered as a relative bias? Welch's t-test~\cite{welch1947generalization} is applied to test the null hypothesis since the sample sizes and variances are usually different between $R_{\Delta}(A=0)$ and $R_{\Delta}(A=1)$. Cohen's d~\cite{cohen2013statistical} is applied to measure the effect size.

\begin{definition}
\textbf{Null hypothesis testing for differential parity.} Given a set of data $X\in\mathbb{R}^d$ with sensitive attribute $A(x\in X)\in\{0, 1\}$, and two sets of decisions on the data $R_{0}(x\in X),\,R_{1}(x\in X) \in \mathbb{R}$, the differential parity t (DPt) score of $R_{0}$ over $R_1$ on $A(X)$ is calculated as \eqref{DPt}.
\begin{equation}
\label{DPt}
\begin{aligned}
DPt(R_{0}, R_{1}, A) 
&= \frac{\overline{R}_{\Delta}(A=1)-\overline{R}_{\Delta}(A=0)}{\sqrt{s^2(\overline{R}_{\Delta}(A=1))+s^2(\overline{R}_{\Delta}(A=0))}}\\
DoF(R_{0}, R_{1}, A) &= \frac{(s^2(\overline{R}_{\Delta}(A=1))+s^2(\overline{R}_{\Delta}(A=0)))^2}{\frac{(s^2(\overline{R}_{\Delta}(A=1)))^2}{|A=1|-1}+\frac{(s^2(\overline{R}_{\Delta}(A=0)))^2}{|A=0|-1}}.
\end{aligned}
\end{equation}
Here $DoF(R_{0}, R_{1}, A)$ is the degrees of freedom measured with Welch's t-test and $s^2(\overline{R}_{\Delta}(A=a)) = \frac{s^2(R_{\Delta}(A=a))}{|A=a|}$ is the sampled variance of $\overline{R}_{\Delta}(A=a)$.

Note that, when the number of test samples is large enough ($|X|>30$), a z-test can be performed with $DPt(R_{0}, R_{1}, A)$ (without $DoF(R_{0}, R_{1}, A)$) instead of the Welch's t-test to simplify the calculation of p value.
\end{definition}

\begin{definition}
\textbf{Effect size for differential parity.} Given a set of data $X\in\mathbb{R}^d$ with sensitive attribute $A(x\in X)\in\{0, 1\}$, and two sets of decisions on the data $R_{0}(x\in X),\,R_{1}(x\in X) \in \mathbb{R}$, the differential parity d (DPd) score of $R_{0}$ over $R_1$ on $A(X)$ is calculated as \eqref{DPd}.
\begin{equation}
\label{DPd}
DPd(R_{0}, R_{1}, A)  = \frac{\overline{R}_{\Delta}(A=1)-\overline{R}_{\Delta}(A=0)}{s}
\end{equation}
where $s=\sqrt{\frac{(|A=1|-1)s^2(R_{\Delta}(A=1))+(|A=0|-1)s^2(R_{\Delta}(A=0))}{|A=1|+|A=0|-2}}$ is the pooled standard deviation.
\end{definition}

\begin{table}[!tbh]
\caption{Effect sizes.}\label{tab:effect_size}
\centering
\small
\begin{tabular}{c|c||c|c|}
\textbf{Effect Size} & \textbf{d} & \textbf{Effect Size} & \textbf{d} \\\midrule
Very Small & 0.01 & Large & 0.8\\  \midrule
Small & 0.2 & Very Large & 1.2\\  \midrule
Medium & 0.5 & Huge & 2.0\\  \bottomrule
\end{tabular}
\end{table}

\noindent\textbf{Evaluation of differential parity:} Utilizing the two metrics, $R_{0}$ is relatively biased towards $A=1$ compared to $R_{1}$ if the null hypothesis is rejected with more than 95\% confidence -- two tailed $p\le0.05$ given the t value $DPt(R_{0}, R_{1}, A)$ and degrees of freedom $DoF(R_{0}, R_{1}, A)$, with at least a small effect size -- $DPd>0.2$ following the magnitude descriptor of Cohen's d in Table~\ref{tab:effect_size}. Figure~\ref{fig:example} shows a toy example of how differential parity is evaluated between two raters on nine face images.


\subsection{Differential Parity Between Decisions on Different Data}
\label{sect:framework}

\textbf{Problem statement: }Given two non-overlapping data $X_0,\,X_1 \in\mathbb{R}^d$ drawn from the same distribution, and two decision sets $R_0(x\in X_0),\, R_1(x\in X_1)\in\mathbb{R}$ made by different entities, evaluate the differential parity between the two decision sets over a binary sensitive attribute $A(x\in X)\in\{0, 1\}$.

\subsubsection{Unbiased Bridge}

The first approach, unbiased bridge, trains a machine learning model $f(x)$ on $(X_0, R_0(x\in X_0))$. Under a naive assumption that $f(x)$ always inherits the bias from its training data, the predictions of $f(x \in X_1)$ will have the same bias as $R_0(x\in X_1)$ and thus can be used to compare against $R_1(x\in X_1)$ for differential parity. As shown in Algorithm~\ref{alg:unbiased_bridge}, $DPt(R_0,R_1,A)$ and $DPd(R_0,R_1,A)$ are estimated as $DPt(f(x\in X_1),R_1(x\in X_1),A)$ and $DPd(f(x\in X_1),R_1(x\in X_1),A)$.

\begin{algorithm}[ht]
\caption{Unbiased Bridge.}\label{alg:unbiased_bridge}
\textbf{Input}: Decisions on one data set \textbf{$(X_0,A(x\in X_0),R_0(x\in X_0))$}.\\
Decisions on another data set \textbf{$(X_1,A(x\in X_1),R_1(x\in X_1))$}.\\
A predictor \textbf{$f(x)$}.\\
\textbf{Output}: Differential parity of $R_0$ over $R_1$ on $A$.\\
\begin{algorithmic}[1] 
        \STATE Fit f(x) on ($X_0,R_0(x\in X_0)$).
        \STATE dpt = DPt(f($X_1$), $R_1(x\in X_1), A)$.
        \STATE dpd = DPd(f($X_1$), $R_1(x\in X_1), A)$.
        \STATE \textbf{return} dpt, dpd
    \end{algorithmic}
\end{algorithm}
The computational complexity of Algorithm~\ref{alg:unbiased_bridge} depends on that of the model training process on $X_0$ in Step 1. Other than that, the calculations of $dpt$ and $dpd$ require time complexity of $O(n)$ where $n=|X_1|$ is the number of test data points.

\subsubsection{Biased Bridge}\label{sec:biased}

Without the naive assumption of $f(x)$ always inheriting every bias in $R_0$, the biased bridge algorithm utilizes both its predictions on $X_0$ and $X_1$ to estimate the differential parity.

\begin{figure*}[!tbh]
  \centering
  \includegraphics[width=\linewidth]{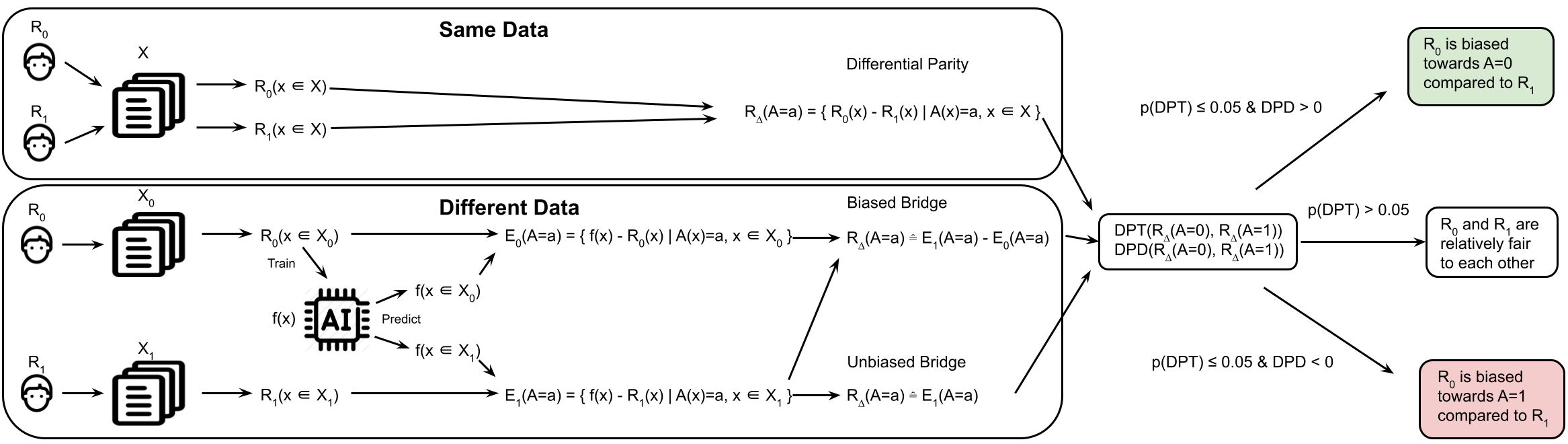}
  \caption{Illustration of how relative fairness is tested with differential parity.}
  \label{fig:dp}
\end{figure*}

\begin{theorem}\label{theorem2}
Under the assumption that the errors of the model predictions $E_{i}(A=a) = \{f(x)-R_i(x) \mid A(x)=a\}$ are i.i.d., the sampled mean and variance of $R_{\Delta}(A=a)$ can be estimated as:
\begin{equation}\label{est1}
\begin{aligned}
\overline{R}_{\Delta}(A=a) 
\hat{=} \overline{E}_{1}(A=a, x\in X_1) - \overline{E}_{0}(A=a, x\in X_0) 
\end{aligned}
\end{equation}
\begin{equation}\label{est2}
\begin{aligned}
s^2(\overline{R}_{\Delta}(A=a)) 
\hat{=} &s^2(\overline{E}_{1}(A=a, x\in X_1))\\ &+s^2(\overline{E}_{0}(A=a, x\in X_0)).
\end{aligned}
\end{equation}
\end{theorem}
\begin{proof}
Under the assumption that the errors of the model predictions are i.i.d., their sampled means should follow normal distribution in large samples: 

\begin{equation}\label{error0}
\begin{aligned}
\frac{\overline{E}_{0}(A=a) - \mu(E_{0}(A=a))}{\sigma(E_{0}(A=a))/\sqrt{|A=a|}} \overset{d}{\rightarrow} \mathcal{N}(0, 1)
\end{aligned}
\end{equation}
\begin{equation}\label{error1}
\begin{aligned}
\frac{\overline{E}_{1}(A=a) - \mu(E_{1}(A=a))}{\sigma(E_{1}(A=a))/\sqrt{|A=a|}} \overset{d}{\rightarrow} \mathcal{N}(0, 1)
\end{aligned}
\end{equation}
where $E_{0}(A=a) = \{f(x)-R_0(x) \mid A(x)=a\}$ and $E_{1}(A=a) = \{f(x)-R_1(x) \mid A(x)=a\}$ are the errors of $f(x)$ compared to $R_0$ and $R_1$.
Given that $R_{\Delta}(A=a) = \{R_{0}(x)-R_{1}(x) \mid A(x)=a\} = E_{1}(A=a) - E_{0}(A=a),$ 
we can estimate:
\begin{equation*}
\begin{aligned}
\overline{R}_{\Delta}(A=a) = &\overline{E}_{1}(A=a) - \overline{E}_{0}(A=a)\\
\hat{=} &\overline{E}_{1}(A=a, x\in X_1) - \overline{E}_{0}(A=a, x\in X_0) \\
s^2(\overline{R}_{\Delta}(A=a)) \hat{=} &s^2(\overline{E}_{1}(A=a, x\in X_1))\\
&+s^2(\overline{E}_{0}(A=a, x\in X_0)).
\end{aligned}
\end{equation*}
\end{proof}

With Theorem~\ref{theorem2}, the differential parity metrics can be calculated as \eqref{DPt} and \eqref{DPd} as shown in Algorithm~\ref{alg:biased_bridge}. The computational complexity of Algorithm~\ref{alg:biased_bridge} depends on that of the model training process on $X_0$ in Step 1. Other than that, the calculations of DPt($R_0$, $R_1$, A) and DPd($R_0$, $R_1$, A) require time complexity of $O(n)$ where $n=|X_0|+|X_1|$ is the total number of data points. 

\begin{algorithm}[!tbh]
\caption{Biased Bridge.}\label{alg:biased_bridge}
\textbf{Input}: Decisions on one data set \textbf{$(X_0,A(x\in X_0),R_0(x\in X_0))$}.\\
Decisions on another data set \textbf{$(X_1,A(x\in X_1),R_1(x\in X_1))$}.\\
A predictor \textbf{$f(x)$}.\\
\textbf{Output}: Differential parity of $R_0$ over $R_1$ on $A$.\\
\begin{algorithmic}[1] 
        \STATE Fit f(x) on ($X_0,R_0(x\in X_0)$).
        \STATE Estimate $\overline{R}_{\Delta}(A=a)$ as \eqref{est1}.
        \STATE Estimate $s^2(\overline{R}_{\Delta}(A=a))$ as \eqref{est2}.
        \STATE \textbf{return} DPt($R_0$, $R_1$, A), DPd($R_0$, $R_1$, A)
    \end{algorithmic}
\end{algorithm}

In summary, Figure~\ref{fig:dp} illustrates how relative fairness is tested with differential parity.

\section{Methodology: Relative Statistical Parity}
\label{sect:methodology2}

\begin{figure}[!tbh]
  \centering
  \includegraphics[width=\linewidth]{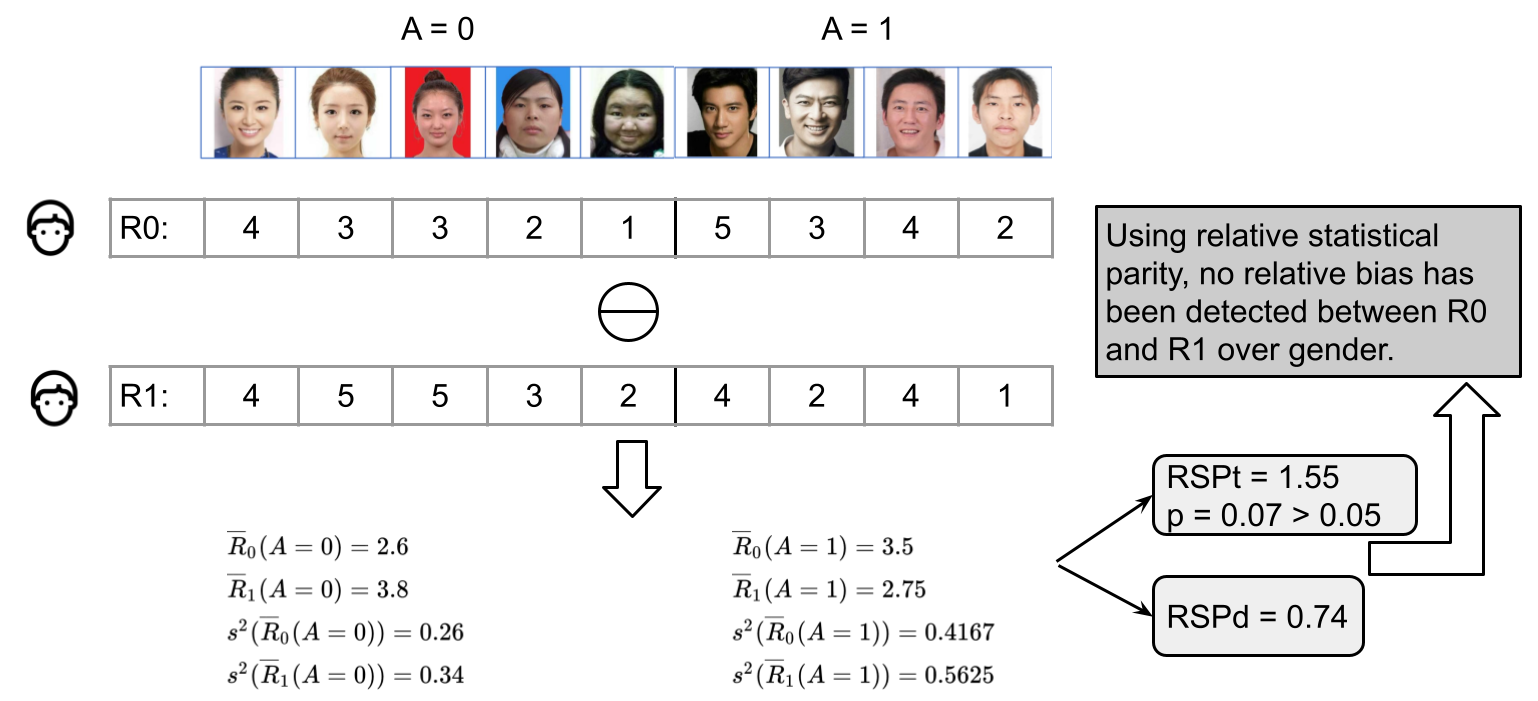}
  \caption{Illustration of relative statistical parity between two raters on face beauty.}
  \label{fig:example_RSP}
\end{figure}

Inspired by statistical parity (also called demographic parity) in group fairness, we can also evaluate relative fairness between two sets of decisions with a novel relative statistical parity. Relative statistical parity requires the \textit{differences between means} of two decision sets are the same across each sensitive group while differential parity requires the \textit{means of differences} between two decision sets are the same across each sensitive group. As a result, differential parity focuses on the individual level of the differences while relative statistical parity measures the group level differences. This leads to a lower statistical power of relative statistical parity than differential parity (similar to the difference in the statistical powers of the unpaired t-tests and paired t-tests). Figure~\ref{fig:example_RSP} illustrates the evaluation of relative statistical parity in an example of face beauty rating. In this example, the same raters were evaluated as not relatively biased against each other when using relative statistical parity because of the lower statistical power of it.

The rest of this section shows the definition of relative statistical parity as well as how to statistically test the violation of relative statistical parity between decisions made either on the same or different data.

\subsection{Relative Statistical Parity}
\label{sect:rsp}

\begin{definition}
\textbf{Relative Statistical Parity.} Two sets of decisions $R_0,\, R_1 \in\mathbb{R}$ satisfy relative statistical parity over the sensitive attribute $A$ if and only if
$$(\overline{R}_0-\overline{R}_1)\perp A.$$
\end{definition}
Unlike differential parity, which focuses on paired decision differences on individual data points, relative statistical parity focuses on differences between means of decisions. From the definitions, it is straightforward that differential parity is a sufficient but not necessary condition for relative statistical parity: $(R_0(x)-R_1(x))\perp A(x) \rightarrow (\overline{R}_0-\overline{R}_1)\perp A$. Therefore, relative statistical parity has lower statistical power than differential parity -- there will be relative biases which can be detected by the violation of differential parity but satisfy relative statistical parity (higher false negative rate or Type II error rate). Our results in Section~\ref{sec:results} will also validate this.

\begin{definition}\label{rsp_binary}
\textbf{Relative Statistical Parity for Binary Sensitive Attributes.} Two sets of decisions $R_0,\, R_1 \in\mathbb{R}$ satisfy relative statistical parity over a binary sensitive attribute $A\in\{0,1\}$ if and only if:
\begin{equation}\label{eq:rsp}
\overline{R}_{0}(A=0)-\overline{R}_{1}(A=0) = \overline{R}_{0}(A=1)-\overline{R}_{1}(A=1).
\end{equation}
\end{definition}
Definition~\ref{rsp_binary} is also equivalent to
$$\overline{R}_{0}(A=0)-\overline{R}_{0}(A=1) = \overline{R}_{1}(A=0)-\overline{R}_{1}(A=1),$$
which requires the statistical parity of $R_0$ and $R_1$ to be the same.

\subsection{Relative Statistical Parity Metrics}
\label{sect:rsp_metrics}

We can also evaluate the violation of relative statistical parity with a null-hypothesis and an effect size test. Unlike differential parity, relative statistical parity can be directly evaluated between decisions made on different data.

\begin{definition}
\textbf{Null hypothesis testing for relative statistical parity.} Given two sets of decisions $R_{0}(x_0\in X_0),\,R_{1}(x_1\in X_1) \in \mathbb{R}$, the t score of the null hypothesis of $R_{0}$ and $R_1$ satisfying relative statistical parity is calculated as \eqref{RSPt}.
\begin{equation}
\label{RSPt}
\begin{aligned}
RSPt(R_{0}, R_{1}, A) 
&= \frac{\overline{R}_{0}(A=1)-\overline{R}_{1}(A=1)-\overline{R}_{0}(A=0)+\overline{R}_{1}(A=0)}{\sqrt{s^2(\overline{R}_{0}(A=1))+s^2(\overline{R}_{1}(A=1))+s^2(\overline{R}_{0}(A=0))+s^2(\overline{R}_{1}(A=0))}}\\
DoF(R_{0}, R_{1}, A) &= \frac{(s^2(\overline{R}_{0}(A=1))+s^2(\overline{R}_{0}(A=0))+s^2(\overline{R}_{1}(A=1))+s^2(\overline{R}_{1}(A=0)))^2}{\frac{(s^2(\overline{R}_{0}(A=1)))^2}{|\{x_0|A(x_0)=1,\,x_0\in X_0\}|-1}+\frac{(s^2(\overline{R}_{0}(A=0)))^2}{|\{x_0|A(x_0)=0,\,x_0\in X_0\}|-1}+\frac{(s^2(\overline{R}_{1}(A=1)))^2}{|\{x_1|A(x_1)=1,\,x_1\in X_1\}|-1}+\frac{(s^2(\overline{R}_{1}(A=0)))^2}{|\{x_1|A(x_1)=0,\,x_1\in X_1\}|-1}},
\end{aligned}
\end{equation}
where
\begin{equation*}
\begin{aligned}
\overline{R}_{i}(A=a) &= \overline{\{R_i(x_i)|A(x_i)=a,\, x_i\in X_i\}}\\
s^2(\overline{R}_{i}(A=a)) &= \frac{s^2(\{R_i(x_i)|A(x_i)=a,\, x_i\in X_i\})}{|\{x_i|A(x_i)=a,\,x_i\in X_i\}|}.
\end{aligned}
\end{equation*}
Similarly, when the number of test samples is large enough ($|X_0|+|X_1|>30$), a z-test can be performed with $RSPt(R_{0}, R_{1}, A)$ (without $DoF(R_{0}, R_{1}, A)$) instead of the Welch's t-test to simplify the calculation of p value.
\end{definition}

\begin{definition}
\textbf{Effect size for relative statistical parity.} Given two sets of decisions $R_{0}(x_0\in X_0),\,R_{1}(x_1\in X_1) \in \mathbb{R}$, the d score of the effect size of $R_{0}$ and $R_1$ violating relative statistical parity is calculated as \eqref{RSPd}.
\begin{equation}
\label{RSPd}
RSPd(R_{0}, R_{1}, A)  = \frac{\overline{R}_{0}(A=1)-\overline{R}_{1}(A=1)-\overline{R}_{0}(A=0)+\overline{R}_{1}(A=0)}{s}
\end{equation}
where $s$ is the pooled standard deviation.
\end{definition}
Note that, $\overline{R}_{\Delta}(A=1)-\overline{R}_{\Delta}(A=0) = \overline{R}_{0}(A=1)-\overline{R}_{1}(A=1)-\overline{R}_{0}(A=0)+\overline{R}_{1}(A=0)$. The nominators of RSPt and RSPd are the same with those of DPt and DPd. Thus the main effect they evaluate is the same, only the variances are different.

In summary, (1) similar to the difference between paired t-tests and unpaired t-tests, relative statistical parity has less statistical power than differential parity; but (2) relative statistical parity can be directly evaluated between decisions made on different data while differential parity has to rely on an external model $f(x)$ to do that.

\begin{table*}[tbh]
\caption{Description of the SCUT-FBP5500 face beauty rating dataset.}
\small
\centering
\setlength\tabcolsep{5pt}
\label{tab:dataset}
\begin{tabular}{c|c|l|c|c|c|c|}
\multirow{2}{*}{Image Size}             & \multirow{2}{*}{\#Images}  & \multicolumn{3}{c|}{Sensitive Attributes $A$}  & \multirow{2}{*}{Beauty Rating $R$} & \multirow{2}{*}{Raters}   \\\cline{3-5}
                    &          &       & {Sex}  & {Race}       &   &  \\\hline
\multirow{2}{*}{$350\times 350$} &    \multirow{2}{*}{$|X|=5,500$}   &  $A=0$ & Female: 2,750    &  Caucasian: 1,500        & \multirow{2}{*}{$R_i(x) \in \{1,2,3,4,5\}$} &  \multirow{2}{*}{$i\in [1, 60]$}    \\
                       &   & $A=1$ & Male: 2,750   &   Asian: 4,000   &       &                    \\\hline
\end{tabular}
\end{table*}
\begin{figure*}[tbh]
  \centering
  \includegraphics[width=0.8\linewidth]{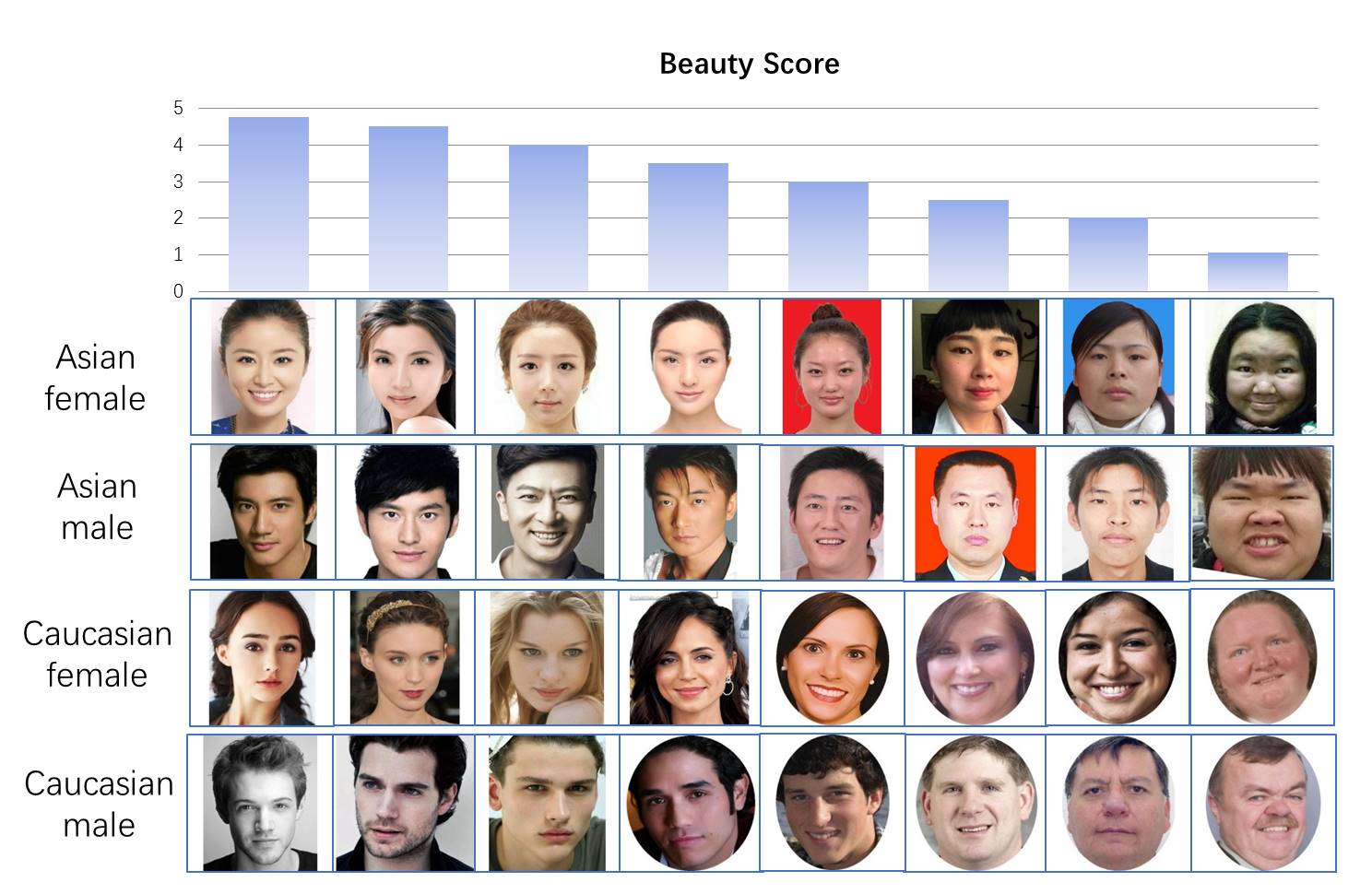}
  \caption{Average beauty scores for some example face images in the SCUT-FBP5500 dataset~\cite{liang2018scut}.}
  \label{fig:scut}
\end{figure*}

\section{Experiment Design}
\label{sec:experiment}

In this section, we design experiments with two real-world case studies\footnote{The source code and data are available at \url{https://github.com/hil-se/Relative-Fairness-Differential-Parity}.} to explore:
\bi
\item
\textbf{RQ1: }How well do the proposed relative fairness notions perform in evaluating relative fairness between two sets of decisions made on the same data?
\item
\textbf{RQ2: }How well do the proposed relative fairness notions perform in evaluating relative fairness between two sets of decisions made on different data?
\item
\textbf{RQ3: }How to use differential parity to analyze relative biases in a real world case study?
\ei
We did not experiment on other popular datasets in machine learning fairness because those datasets only offer one set of ground truth labels. Our experiment requires human decisions from different raters and the SCUT-FBP5500 dataset~\cite{liang2018scut} was the only public data we could find with labels from different raters. The Ph.D. admission data was collected by the authors and it also provides independent decisions from different committee members (thus their decisions on each applicant should be i.i.d.). The sample size of the Ph.D. admission data is relatively smaller but still is larger than 30 and therefore the central limit theorem applies.

\subsection{Case Study on Face Beauty Rating}\label{sect:casestudy}

As shown in Table~\ref{tab:dataset}, we select the SCUT-FBP5500 dataset~\cite{liang2018scut} as our first case study to explore RQ1 and RQ2. This dataset has 5,500 face images, each of the face image has two sensitive attributes -- Sex and Race, and each image has been rated with a 1 (least beautiful) to 5 (most beautiful) subjective beauty score by the same 60 human raters. This is the only dataset we found which has both sensitive attributes and decisions from different human annotators. The only disadvantage of this dataset is that, there is usually no ethical issue in having biased ratings of the beauty scores (e.g. there is nothing to blame if one rater prefers Asian Male over Caucasian Male.) But this is also the reason why this dataset can be available online to safely analyze the bias from different human raters. 

In this case study, we utilized the ratings from the first three human raters ($R_1$, $R_2$, and $R_3$) and the average ratings of the 60 humans ($R_{Avg}$) as four different decision sets. Figure~\ref{fig:scut} shows the average ratings for some example face images.

\subsubsection{RQ1: Relative fairness on the same data}\label{sect:RQ1}

RQ1 evaluates the performance of the proposed two relative fairness notions between two sets of decisions made on the same data. First, differential parity and relative statistical parity are evaluated between pairs of ratings $(R_i,R_j),\quad i,j\in \{1,2,3,Avg\}$ on all images. These results serve as the ground truth evaluations. Next, 50\% of the images are randomly sampled to evaluate these two fairness notions and these evaluations are repeated for 10,000 times. Type I error rates and Type II error rates of differential parity and relative statistical parity can thus be analyzed by comparing the evaluation results on 50\% random samples and the ground truths.

\subsubsection{RQ2: Relative fairness on different data}\label{sect:RQ2}

RQ2 evaluates the performance of the proposed two relative fairness notions between two sets of decisions made on different data. Each time, the image set is randomly split into 50\% as $X_i$ and 50\% as $X_j$. For a pair of raters $(R_i,R_j)$, only $R_i(x_i\in X_i)$ and  $R_j(x_j\in X_j)$ are known.

For relative statistical parity, $RSPt$ and $RSPd$ can be directly calculated with $R_i(x_i\in X_i)$ and $R_j(x_j\in X_j)$.

For differential parity, a machine learning model $f(x)$ is trained on $(X_i, R_i(x_i\in X_i))$. We have tested multiple deep neural network architectures including ResNet-50, VGG-16, and AlexNet for $f(x)$. Among them, the VGG-16 model~\cite{simonyan2014very} with pre-trained weights on the ImageNet data, and the last four layers being replaced by a dense layer of size 256 and a one-node linear output layer achieved the best performance in terms of Pearson correlation coefficient in predicting the ratings of $R_i$ on data $X_j$. Using this model $f(x)$, unbiased bridge and biased bridge algorithms are applied to approximate the differential parity metrics between $R_i(x_i\in X_i)$ and  $R_j(x_j\in X_j)$.

\begin{table*}[!tbh]
\caption{Description of the Ph.D. admission dataset.}
\small
\centering
\setlength\tabcolsep{5pt}
\label{tab:phd}
\begin{tabular}{l|c|c|c|c|c|}
 Year & \#Features             &\#Applicants  & {Sensitive Attribute $A$} &Ratings $R$ & Raters    \\\hline
\multirow{2}{*}{2022}         &     \multirow{2}{*}{21}            &       \multirow{2}{*}{148}   & $A=0$: No US degree        &   \multirow{2}{*}{$R_i(x)\in [1,7]$} & \multirow{2}{*}{$i\in \{1,2,3,4,5,\text{Committee}\}$} \\
&&& $A=1$: Has US degree && \\\hline
        \multirow{2}{*}{2023}         &     \multirow{2}{*}{21}            &       \multirow{2}{*}{174}   & $A=0$: No US degree        &   \multirow{2}{*}{$R_i(x)\in [1,7]$} & \multirow{2}{*}{$i\in \{2,3,4,5,6,\text{Committee}\}$} \\
&&& $A=1$: Has US degree && \\\hline
\end{tabular}
\end{table*}

\subsection{Case Study on Ph.D. Admission}\label{sect:casestudy2}

In RQ3, we use the Ph.D. admission data shown in Table~\ref{tab:phd} to demonstrate a real application scenario of differential parity. This dataset was collected by the first author from the Computing and Information Sciences Ph.D. program of Rochester Institute of Technology in two consecutive years (2022-2023). The dataset includes partial student application information such as GRE and GPA. It also has human evaluation results from five admission committee members each year. From 2022 to 2023, four members stayed in the committee while one member ($R_1$) was replaced by a new member ($R_6$). Each year, each member of the admission committee first independently rated how qualified an applicant was on a scale from $1$ to $7$ based on the complete application information, including recommendation letters, personal statements, and CVs. The five members then met to discuss and come to a consensus rating $R_{\text{Committee}}$. Due to privacy concerns, sensitive/PII information such as race, gender, CV, personal statement, or recommendation letters are not included in the data. We use the attribute ``US Degree'' as the sensitive attribute $A$ for bias analysis--- whether a committee member has preference over applicants owning a US degree over those who don't. Despite its small size and partial information, this dataset reflects a real application scenario of this work and is thus being explored to answer RQ3.

\subsubsection{RQ3: Relative bias in Ph.D. admission}\label{sect:RQ3}

As described in Table~\ref{tab:scenario}, there are many types of relative biases we can evaluate using differential parity. In this Ph.D. admission case study, we will demonstrate how differential parity could be used to improve fairness in the Ph.D. admission process.
\bi
\item
Within each year, the relative biases between each committee member's ratings can be directly evaluated with differential parity. This resembles Scenario 4 of Table~\ref{tab:scenario}. 
\item
Within each year, $R_{\text{Committee}}$ could be treated as the ground truth since it represents the final consensus decision of the committee. The relative biases between each committee member's ratings and $R_{\text{Committee}}$ can be evaluated with differential parity. This resembles Scenario 2 of Table~\ref{tab:scenario} and shows whether some of the committee members were initially biased.
\item
Differential parity between decisions made in 2022 and 2023 can also be evaluated with the biased bridge algorithm. In this case, a support vector regressor is trained on ratings from 2022 $R_i(x_i\in X_{2022})$. Its predictions are utilized to calculate the differential parity metrics between $R_i(x_i\in X_{2022})$ and $R_j(x_j\in X_{2023})$. Relative bias between $R_1$ and $R_6$ can only be evaluated using biased bridge since they have never made decisions on the same data. In addition, the relative bias between ratings from the same rater $R_i$ in 2022 and 2023 can also be evaluated to see whether a committee member has changed in his or her way of rating the applicants. This resembles Scenario 5 of Table~\ref{tab:scenario}.
\item
Another realistic application is the evaluation of differential parity between $R_{\text{Committee}}$ in 2022 and $R_j$ in 2023. This could happen after each committee member has completed the ratings and before the committee reaches a consensus. Using previous year's ground truth, each committee member can test whether his or her ratings have unwanted biases. This could also facilitate the discussion of $R_{\text{Committee}}$ in 2023. This resembles Scenario 3 of Table~\ref{tab:scenario}.
\ei

\subsection{Execution Environment}\label{sect:environment}

All the experiments were executed on 4 NVIDIA A100 Tensor Core GPUs with 128GB RAM on RIT's research computing services~\cite{https://doi.org/10.34788/0s3g-qd15}. We used a random seed of $0$ throughout the experiments. However, due to unknown memory leak problems, results in Table~\ref{tab:rq2m} were generated by splitting into multiple runs. So, the results of Table~\ref{tab:rq2m} by running the provided code at \url{https://github.com/hil-se/RelativeFairnessTesting} could be slightly different.  The VGG-16 model used in RQ2 is implemented with Tensorflow Keras, it optimizes for Huber loss with stochastic gradient descent. Twenty percent of the training data is used as validation, the maximum number of epochs is set to $50$, and the batch size is $10$. The support vector regression model used in RQ3 is implemented with scikit-learn with an RBF kernel and the default parameters.

\begin{table*}[tbh]
\caption{Results on two sensitive attributes Sex and Race are shown with numbers of (p value) and d value for the ground truth differential parity and relative statistical parity. Detected violations are colored as \colorbox{green!20}{green} if $R_i$ prefers $A=1$ more than $R_j$ ($p<0.05$ and $d\ge 0.2$) or colored as \colorbox{red!20}{red} if $R_i$ prefers $A=0$ more than $R_j$ ($p<0.05$ and $d\le 0.2$). Each sampled result is repeated $10,000$ times and the percentage of detecting a violation ($p<0.05$ and $|d|\ge 0.2$) is reported.}
\centering
\small
\setlength\tabcolsep{1.5pt}
\label{tab:rq1}
\begin{tabular}{l|c|rr|rr|rr|rr|}
\multicolumn{1}{c}{}                 & \multirow{2}{*}{\diagbox{$R_i$}{$R_j$}} & \multicolumn{1}{c|}{Sex} & \multicolumn{1}{c|}{Race} & \multicolumn{1}{c|}{Sex} & \multicolumn{1}{c|}{Race} & \multicolumn{1}{c|}{Sex} & \multicolumn{1}{c|}{Race} & \multicolumn{1}{c|}{Sex} & \multicolumn{1}{c|}{Race} \\ \cline{3-10} 
\multicolumn{1}{c}{}    &                                                         & \multicolumn{2}{c|}{$R_1$}                           & \multicolumn{2}{c|}{$R_2$}                           & \multicolumn{2}{c|}{$R_3$}                           & \multicolumn{2}{c|}{$R_{Avg}$}                       \\ \hline
{Ground truth DP}  & \multirow{4}{*}{$R_1$}                                     & (1.00) 0.00              & (1.00) 0.00               & \cellcolor{red!20} (0.00) -0.44             & \cellcolor{red!20} (0.00) -0.24              & \cellcolor{red!20} (0.00) -0.49             & (0.01) -0.09              & \cellcolor{red!20} (0.00) -0.47             & \cellcolor{red!20} (0.00) -0.24              \\
{Ground truth RSP} &                                           & (1.00) 0.00              & (1.00) 0.00               &  \cellcolor{red!20} (0.00) -0.22             & (0.00) -0.12              & \cellcolor{red!20} (0.00) -0.20             & (0.09) -0.04              & (0.00) -0.19             & (0.00) -0.10              \\
{Sampled DP}       &                                           & 0                        & 0                         & 1                        & 0.9                       & 1                        & 0                         & 1                        & 0.94                      \\
{Sampled RSP}      &                                           & 0                        & 0                         & 0.94                     & 0                         & 0.51                     & 0                         & 0.1                      & 0                         \\ \hline
{Ground truth DP}  & \multirow{4}{*}{$R_2$}                                     & \cellcolor{green!20} (0.00) 0.44              & \cellcolor{green!20} (0.00) 0.24               & (1.00) 0.00              & (1.00) 0.00               & (0.04) -0.06             & (0.00) 0.15               & (0.00) 0.13              & (0.00) 0.08               \\
{Ground truth RSP} &                                           & \cellcolor{green!20} (0.00) 0.22              & (0.00) 0.12               & (1.00) 0.00              & (1.00) 0.00               & (0.15) -0.03             & (0.00) 0.08               & (0.02) 0.04              & (0.18) 0.03               \\
{Sampled DP}       &                                           & 1                        & 0.88                      & 0                        & 0                         & 0                        & 0.04                      & 0                        & 0                         \\
{Sampled RSP}      &                                           & 0.95                     & 0                         & 0                        & 0                         & 0                        & 0                         & 0                        & 0                         \\ \hline
{Ground truth DP}  & \multirow{4}{*}{$R_3$}                                     & \cellcolor{green!20} (0.00) 0.49              & (0.01) 0.09               & (0.04) 0.06              & (0.00) -0.15              & (1.00) 0.00              & (1.00) 0.00               & (0.00) 0.15              & (0.00) -0.14              \\
{Ground truth RSP} &                                           & \cellcolor{green!20} (0.00) 0.20              & (0.09) 0.04               & (0.15) 0.03              & (0.00) -0.08              & (1.00) 0.00              & (1.00) 0.00               & (0.00) 0.06              & (0.01) -0.05              \\
{Sampled DP}       &                                           & 1                        & 0                         & 0                        & 0.04                      & 0                        & 0                         & 0.04                     & 0.04                      \\
{Sampled RSP}      &                                           & 0.49                     & 0                         & 0                        & 0                         & 0                        & 0                         & 0                        & 0                         \\ \hline
{Ground truth DP}  & \multirow{4}{*}{$R_{Avg}$}                                 & \cellcolor{green!20} (0.00) 0.47              & \cellcolor{green!20} (0.00) 0.24               & (0.00) -0.13             & (0.00) -0.08              & (0.00) -0.15             & (0.00) 0.14               & (1.00) 0.00              & (1.00) 0.00               \\ 
{Ground truth RSP} &                                           & (0.00) 0.19              & (0.00) 0.10               & (0.02) -0.04             & (0.18) -0.03              & (0.00) -0.06             & (0.01) 0.05               & (1.00) 0.00              & (1.00) 0.00               \\
{Sampled DP}       &                                           & 1                        & 0.94                      & 0                        & 0                         & 0.03                     & 0.03                      & 0                        & 0                         \\
{Sampled RSP}      &                                           & 0.11                     & 0                         & 0                        & 0                         & 0                        & 0                         & 0                        & 0   \\\hline                     
\end{tabular}

\end{table*}

\section{Experimental Results}
\label{sec:results}

\noindent \textbf{RQ1: How well do the proposed relative fairness notions perform in evaluating relative fairness between two sets of decisions made on the same data?} 

Table~\ref{tab:rq1} shows the experimental results on the SCUT-FBP5500 dataset. In this table, two relative fairness notions differential parity (DP) and relative statistical parity (RSP) are evaluated between a pair of decisions $(R_i, R_j)$. The four row indices are explained below:
\bi
\item
\textit{Ground Truth DP}: differential parity evaluated between $R_i(x\in X)$ and $R_j(x\in X)$, where $X$ is the set of all face images from the SCUT-FBP5500 dataset. The result shows the p value of the null hypothesis test in the brackets and the d value of the effect size $DPd(R_i, R_j, A)$ with its sign representing the direction of the relative bias.
\item
\textit{Ground Truth RSP}: relative statistical parity evaluated between $R_i(x\in X)$ and $R_j(x\in X)$. The result shows the p value of the null hypothesis test in the brackets and the d value of the effect size $RSPd(R_i, R_j, A)$ with its sign representing the direction of the relative bias.
\item
\textit{Sampled DP}: differential parity evaluated between $R_i(x\in X_s)$ and $R_j(x\in X_s)$, where $X_s\subset X$ consists $50\%$ of randomly sampled images from the SCUT-FBP5500 dataset without replacement. This process is repeated $10,000$ times and the frequency of $R_i$ and $R_j$ violating differential parity on the sampled data $X_s$ is shown in the table.
\item
\textit{Sampled RSP}: relative statistical parity evaluated between $R_i(x\in X_s)$ and $R_j(x\in X_s)$, where $X_s\subset X$ consists $50\%$ of randomly sampled images from the SCUT-FBP5500 dataset without replacement.This process is repeated $10,000$ times and the frequency of $R_i$ and $R_j$ violating differential parity on the sampled data $X_s$ is shown in the table.
\ei

From Table~\ref{tab:rq1} we can observe:
\be
\item
All the evaluation results are symmetric between $(R_i, R_j)$ and $(R_j, R_i)$. This validates our previous claim of the proposed relative fairness notions. This is an advantage of differential parity and relative statistical parity over existing notions such as separation and sufficiency.
\item
Comparing ground truth DP and ground truth RSP, we can see that differential parity does have higher statistical power than relative statistical parity as discussed in Section~\ref{sect:rsp}. There are three cases when differential parity confirmed a significant relative bias but relative statistical parity did not.
\item
The sampled results also confirm that differential parity has higher statistical power than relative statistical parity. With half the sample size, differential parity still reliably detects the relative biases in higher than 88\% true positive rates and lower than $4\%$ false positive rates. On the other hand, relative statistical parity between $R_1$ and $R_3$ on Sex only has $50\%$ true positive rate.
\item
Based on the ground truth DP results, we can conclude that $R_1$ overrates Female ($A=0$) and underrates Male ($A=1$) compared to all other raters. $R_1$ also overrates Caucasian ($A=0$) and underrates Asian ($A=1$) compared to $R_2$ and $R_{Avg}$. The other three raters are relatively fair against each other.
\ee

\noindent\textbf{Answer to RQ1.} Both differential parity and relative statistical parity are symmetric between $(R_i, R_j)$ and $(R_j, R_i)$. Differential parity has higher statistical power than relative statistical parity in detecting relative biases between decision sets.

\begin{table*}[tbh]
\caption{Results on two sensitive attributes Sex and Race are shown with numbers of (p value) and d value for the ground truth differential parity and relative statistical parity. Detected violations are colored as \colorbox{green!20}{green} if $R_i$ prefers $A=1$ more than $R_j$ ($p<0.05$ and $d\ge 0.2$) or colored as \colorbox{red!20}{red} if $R_i$ prefers $A=0$ more than $R_j$ ($p<0.05$ and $d\le 0.2$).}
\centering
\small
\setlength\tabcolsep{1.5pt}
\label{tab:rq2}
\begin{tabular}{l|c|rr|rr|rr|rr|}
\multicolumn{1}{c}{}                 & \multirow{2}{*}{\diagbox{$R_i$}{$R_j$}} & \multicolumn{1}{c|}{Sex} & \multicolumn{1}{c|}{Race} & \multicolumn{1}{c|}{Sex} & \multicolumn{1}{c|}{Race} & \multicolumn{1}{c|}{Sex} & \multicolumn{1}{c|}{Race} & \multicolumn{1}{c|}{Sex} & \multicolumn{1}{c|}{Race} \\ \cline{3-10} 
\multicolumn{1}{c}{}    &                                                         & \multicolumn{2}{c|}{$R_1$}                           & \multicolumn{2}{c|}{$R_2$}                           & \multicolumn{2}{c|}{$R_3$}                           & \multicolumn{2}{c|}{$R_{Avg}$}                       \\ \hline
{Ground truth DP}    & \multirow{5}{*}{$R_1$}                & (1.00) 0.00                      & {(1.00) 0.00}  & \cellcolor{red!20} (0.00) -0.44             & \cellcolor{red!20} {(0.00) -0.24} & \cellcolor{red!20} (0.00) -0.49             & (0.01) -0.09              & \cellcolor{red!20} (0.00) -0.47             & \cellcolor{red!20} (0.00) -0.24              \\
{Ground truth RSP}   &                       & (1.00) 0.00                      & {(1.00) 0.00}  & \cellcolor{red!20} (0.00) -0.22             & {(0.00) -0.12} & \cellcolor{red!20} (0.00) -0.20             & (0.09) -0.04              & (0.00) -0.19             & (0.00) -0.10              \\
{DP unbiased bridge} &                       & (0.01) -0.10                     & {(0.13) -0.07} & \cellcolor{red!20} (0.00) -0.63             & \cellcolor{red!20} {(0.00) -0.35} & \cellcolor{red!20} (0.00) -0.60             & (0.00) -0.14              &  \cellcolor{red!20} (0.00) -0.73             & \cellcolor{red!20} (0.00) -0.37              \\
{DP biased bridge}   &                       & (0.43) 0.01                      & {(0.16) -0.04} & \cellcolor{red!20} (0.00) -0.45             & \cellcolor{red!20} {(0.00) -0.28} & \cellcolor{red!20} (0.00) -0.45             & (0.01) -0.11              & \cellcolor{red!20} (0.00) -0.46             & \cellcolor{red!20} (0.00) -0.27              \\
{RSP}                &  & (0.34) 0.03                      & {(0.92) -0.00} & \cellcolor{red!20} (0.00) -0.20             & {(0.00) -0.13} & (0.00) -0.17             & (0.26) -0.03              & (0.00) -0.16             & (0.00) -0.10              \\ \hline
{Ground truth DP}    & \multirow{5}{*}{$R_2$}                  & \cellcolor{green!20} (0.00) 0.44                      & \cellcolor{green!20} {(0.00) 0.24} & (1.00) 0.00              & {(1.00) 0.00}  & (0.04) -0.06             & (0.00) 0.15               & (0.00) 0.13              & (0.00) 0.08               \\
{Ground truth RSP}   &                       & \cellcolor{green!20} (0.00) 0.22                      & {(0.00) 0.12} & (1.00) 0.00              & {(1.00) 0.00}  & (0.15) -0.03             & (0.00) 0.08               & (0.02) 0.04              & (0.18) 0.03               \\
{DP unbiased bridge} &                       & \cellcolor{green!20} (0.00) 0.39                      & \cellcolor{green!20} {(0.00) 0.20} & (0.36) -0.03             & {(0.19) -0.06} & (0.01) -0.10             & (0.00) 0.13               & (0.10) 0.06              & (0.89) 0.01               \\
{DP biased bridge}   &                       & \cellcolor{green!20} (0.00) 0.41                      & \cellcolor{green!20} {(0.00) 0.23} & (0.49) 0.00              & {(0.35) 0.02}  & (0.02) -0.08             & (0.00) 0.16               & (0.01) 0.09              & (0.03) 0.07               \\
{RSP}                &  & \cellcolor{green!20} (0.00) 0.24                      & {(0.00) 0.13} & (0.20) 0.03              & {(0.45) 0.02}  & (0.49) -0.02             & (0.00) 0.10               & (0.01) 0.07              & (0.16) 0.04               \\ \hline
{Ground truth DP}    & \multirow{5}{*}{$R_3$}                & \cellcolor{green!20} (0.00) 0.49                      & {(0.01) 0.09} & (0.04) 0.06              & {(0.00) -0.15} & (1.00) 0.00              & (1.00) 0.00               & (0.00) 0.15              & (0.00) -0.14              \\
{Ground truth RSP}   &                       & \cellcolor{green!20} (0.00) 0.20                      & {(0.09) 0.04} & (0.15) 0.03              & {(0.00) -0.08} & (1.00) 0.00              & (1.00) 0.00               & (0.00) 0.06              & (0.01) -0.05              \\
{DP unbiased bridge} &                       & \cellcolor{green!20} (0.00) 0.41                      & {(0.05) 0.09} & (0.08) -0.07             & \cellcolor{red!20} {(0.00) -0.21} & (0.00) -0.16             & (0.86) 0.01               & (0.91) -0.00             & \cellcolor{red!20} (0.00) -0.22              \\
{DP biased bridge}   &                       & \cellcolor{green!20} (0.00) 0.39                      & {(0.00) 0.11} & (0.09) -0.05             & {(0.00) -0.15} & (0.00) -0.14             & (0.15) 0.04               & (0.42) 0.01              & (0.00) -0.14              \\
{RSP}                &  & \cellcolor{green!20} {(0.00) 0.20} & {(0.09) 0.05} & (0.18) 0.04              & {(0.07) -0.05} & (0.88) -0.00             & (0.49) 0.02               & (0.03) 0.06              & (0.19) -0.04              \\ \hline
{Ground truth DP}    & \multirow{5}{*}{$R_{Avg}$}             & \cellcolor{green!20} (0.00) 0.47                      & \cellcolor{green!20} {(0.00) 0.24} & (0.00) -0.13             & {(0.00) -0.08} & (0.00) -0.15             & (0.00) 0.14               & (1.00) 0.00              & (1.00) 0.00               \\
{Ground truth RSP}   &                       & (0.00) 0.19                      & {(0.00) 0.10} & (0.02) -0.04             & {(0.18) -0.03} & (0.00) -0.06             & (0.01) 0.05               & (1.00) 0.00              & (1.00) 0.00               \\
{DP unbiased bridge} &                       & \cellcolor{green!20} (0.00) 0.38                      & \cellcolor{green!20} {(0.00) 0.22} & (0.00) -0.16             & {(0.12) -0.06} & (0.00) -0.19             & (0.00) 0.14               & (0.01) -0.10             & (0.78) -0.01              \\
{DP biased bridge}   &                       & \cellcolor{green!20} (0.00) 0.37                      & \cellcolor{green!20} {(0.00) 0.24} & (0.00) -0.18             & {(0.35) -0.02} & (0.00) -0.20             & (0.00) 0.17               & (0.00) -0.13             & (0.09) 0.06               \\
{RSP}                &  & (0.00) 0.19                      & (0.00) 0.11                      & (0.42) -0.02             & (0.64) -0.01                      & (0.04) -0.06             & (0.02) 0.07               & (0.69) 0.01              & (0.85) 0.01             \\\hline 
\end{tabular}

\end{table*}

\noindent \textbf{RQ2: How well do the proposed relative fairness notions perform in evaluating relative fairness between two sets of decisions made on different data?} 

Table~\ref{tab:rq2} presents the results of a one-time test between decisions made on different data. The five row indices are explained below:
\bi
\item
\textit{Ground Truth DP}: same as in Table~\ref{tab:rq1}.
\item
\textit{Ground Truth RSP}: same as in Table~\ref{tab:rq1}.
\item
\textit{DP unbiased bridge}: differential parity evaluated between $R_i(x_i\in X_i)$ and $R_j(x_j\in X_j)$, where $X_i \cup X_j = X$ and $|X_i|=|X_j|=\frac{1}{2}|X|$. With unbiased bridge, a model $f(x)$ is trained on $(X_i, R_i(x_i\in X_i))$ and its predictions $f(x_j\in X_j)$ are tested against $R_j(x_j\in X_j)$.
\item
\textit{DP biased bridge}: differential parity evaluated between $R_i(x_i\in X_i)$ and $R_j(x_j\in X_j)$. With biased bridge, a model $f(x)$ is trained on $(X_i, R_i(x_i\in X_i))$ and its predictions $f(x_i\in X_i)$ and $f(x_j\in X_j)$ are used to evaluate differential parity between $R_i$ and $R_j$.
\item
\textit{RSP}: relative statistical parity evaluated between $R_i(x_i\in X_i)$ and $R_j(x_j\in X_j)$.
\ei

From Table~\ref{tab:rq2} we can observe:
\be
\item
Although the decisions were made on different data, the RSP evaluated between $R_i(x_i\in X_i)$ and $R_j(x_j\in X_j)$ is very consistent with the ground truth RSP evaluated between $R_i(x\in X)$ and $R_j(x\in X)$. This shows that relative statistical parity works perfectly between decisions made on different data. However, it still has lower statistical power than differential parity.
\item
The biased bridge results are always consistent with the ground truth DP results. On the other hand, there are two cases where the unbiased bridge detects a violation of differential parity but the ground truth DP does not. Therefore, in the evaluation of differential parity between decisions made on different data, biased bridge has lower error rate than unbiased bridge. The effect size results (DPd) evaluated by biased bridge are also closer to the ground truth DP evaluations (with an average error of $0.043$) than those evaluated by unbiased bridge (with an average error of $0.074$).
\item
When the ground truth DP results are considered the ground truths, DP biased bridge results has $0$ error rate and performs better than RSP.
\ee

Since Table~\ref{tab:rq2} only shows a one-time test result, the conclusions could be largely affected by the random split of $X_i$ and $X_j$. So we repeated the experiment in Table~\ref{tab:rq2} 20 times with different random splits of $X_i$ and $X_j$. The corresponding results are shown in Table~\ref{tab:rq2m}, where the percentage of times a relative bias is detected ($p<0.05$ and $|d|\ge 0.2$) is shown for the DP unbiased bridge, DP biased bridge, and RSP rows. Using Table~\ref{tab:rq2m}, we can better analyze the Type I and Type II error rates of each treatment by treating the ground truth DP results as the ground truths -- (1) when a ground truth DP result is not highlighted by colors, no relative bias should be detected, so the corresponding number of a treatment would be its false positive rate (Type I error rate); when a ground truth DP result is highlighted by colors, a relative bias should be detected, so the corresponding number of a treatment would be its true positive rate ($1-$Type II error rate). Table~\ref{tab:error} shows the summarized results of Type I and Type II error rates of the three treatments. We can see that the biased bridge algorithm achieved lower Type I and Type II error rates than the unbiased bridge algorithm. Although RSP has $0$ Type I error rate, its Type II error rate is much higher than DP-based approaches. Therefore, our conclusion on RQ2 stays the same -- biased bridge best evaluates the relative fairness between decisions made on different data, with both Type I and Type II error rates lower than $0.1$.

\begin{table*}[tbh]
\caption{Results on two sensitive attributes Sex and Race are shown with numbers of (p value) and d value for the ground truth differential parity and relative statistical parity. Detected violations are colored as \colorbox{green!20}{green} if $R_i$ prefers $A=1$ more than $R_j$ ($p<0.05$ and $d\ge 0.2$) or colored as \colorbox{red!20}{red} if $R_i$ prefers $A=0$ more than $R_j$ ($p<0.05$ and $d\le 0.2$). Each sampled result is repeated $20$ times and the percentage of detecting a relative bias ($p<0.05$ and $|d|\ge 0.2$) is reported.}
\centering
\small
\setlength\tabcolsep{1.5pt}
\label{tab:rq2m}
\begin{tabular}{l|c|ll|ll|ll|ll|}
\multicolumn{1}{c}{} & \multirow{2}{*}{\diagbox{$R_i$}{$R_j$}} & \multicolumn{1}{c|}{Sex} & \multicolumn{1}{c|}{Race} & \multicolumn{1}{c|}{Sex} & \multicolumn{1}{c|}{Race} & \multicolumn{1}{c|}{Sex} & \multicolumn{1}{c|}{Race} & \multicolumn{1}{c|}{Sex} & \multicolumn{1}{c|}{Race} \\ \cline{3-10} 
\multicolumn{1}{c}{} &                                         & \multicolumn{2}{c|}{$R_1$}                           & \multicolumn{2}{c|}{$R_2$}                           & \multicolumn{2}{c|}{$R_3$}                           & \multicolumn{2}{c|}{$R_{Avg}$}                       \\ \hline
{Ground truth DP}    & \multirow{5}{*}{$R_1$}                & (1.00) 0.00                      & {(1.00) 0.00}  & \cellcolor{red!20} (0.00) -0.44             & \cellcolor{red!20} {(0.00) -0.24} & \cellcolor{red!20} (0.00) -0.49             & (0.01) -0.09              & \cellcolor{red!20} (0.00) -0.47             & \cellcolor{red!20} (0.00) -0.24              \\
{Ground truth RSP}   &                       & (1.00) 0.00                      & {(1.00) 0.00}  & \cellcolor{red!20} (0.00) -0.22             & {(0.00) -0.12} & \cellcolor{red!20} (0.00) -0.20             & (0.09) -0.04              & (0.00) -0.19             & (0.00) -0.10              \\
DP unbiased bridge    &                                         & 0.05                     & 0.05                      & 1                        & 0.65                      & 1                        & 0.15                      & 1                        & 0.8                       \\
DP biased bridge      &                                         & 0                        & 0                         & 1                        & 0.85                      & 1                        & 0                         & 1                        & 0.7                       \\
RSP                   &                                         & 0                        & 0                         & 0.85                     & 0                         & 0.6                      & 0                         & 0.4                      & 0                         \\ \hline
Ground truth DP       & \multirow{5}{*}{$R_2$}                  & \cellcolor{green!20} (0.00) 0.44              & \cellcolor{green!20} (0.00) 0.24               & (1.00) 0.00              & (1.00) 0.00               & (0.04) -0.06             & (0.00) 0.15               & (0.00) 0.13              & (0.00) 0.08               \\
Ground truth RSP      &                                         & \cellcolor{green!20} (0.00) 0.22              & (0.00) 0.12               & (1.00) 0.00              & (1.00) 0.00               & (0.15) -0.03             & (0.00) 0.08               & (0.02) 0.04              & (0.18) 0.03               \\
DP unbiased bridge    &                                         & 1                        & 0.85                      & 0.05                     & 0                         & 0                        & 0.05                      & 0.05                     & 0.1                       \\
DP biased bridge      &                                         & 1                        & 0.8                       & 0                        & 0                         & 0                        & 0                         & 0                        & 0                         \\
RSP                   &                                         & 0.9                      & 0                         & 0                        & 0                         & 0                        & 0                         & 0                        & 0                         \\ \hline
Ground truth DP       & \multirow{5}{*}{$R_3$}                  & \cellcolor{green!20} (0.00) 0.49              & (0.01) 0.09               & (0.04) 0.06              & (0.00) -0.15              & (1.00) 0.00              & (1.00) 0.00               & (0.00) 0.15              & (0.00) -0.14              \\
Ground truth RSP      &                                         & \cellcolor{green!20} (0.00) 0.20              & (0.09) 0.04               & (0.15) 0.03              & (0.00) -0.08              & (1.00) 0.00              & (1.00) 0.00               & (0.00) 0.06              & (0.01) -0.05              \\
DP unbiased bridge    &                                         & 1                        & 0.2                       & 0.1                      & 0.45                      & 0.3                      & 0.1                       & 0.1                      & 0.35                      \\
DP biased bridge      &                                         & 1                        & 0                         & 0                        & 0.1                       & 0.1                      & 0                         & 0                        & 0.05                      \\
RSP                   &                                         & 0.5                      & 0                         & 0                        & 0                         & 0                        & 0                         & 0                        & 0                         \\ \hline
Ground truth DP       & \multirow{5}{*}{$R_{Avg}$}              & \cellcolor{green!20} (0.00) 0.47              & \cellcolor{green!20} (0.00) 0.24               & (0.00) -0.13             & (0.00) -0.08              & (0.00) -0.15             & (0.00) 0.14               & (1.00) 0.00              & (1.00) 0.00               \\
Ground truth RSP      &                                         & (0.00) 0.19              & (0.00) 0.10               & (0.02) -0.04             & (0.18) -0.03              & (0.00) -0.06             & (0.01) 0.05               & (1.00) 0.00              & (1.00) 0.00               \\
DP unbiased bridge    &                                         & 1                        & 0.8                       & 0.6                      & 0                         & 0.25                     & 0.1                       & 0.35                     & 0                         \\
DP biased bridge      &                                         & 1                        & 0.9                       & 0.3                      & 0                         & 0.25                     & 0.1                       & 0.05                     & 0                         \\
RSP                   &                                         & 0.35                     & 0                         & 0                        & 0                         & 0                        & 0                         & 0                        & 0                         \\ \hline
\end{tabular}

\end{table*}

\begin{table}[!tbh]
\caption{Summarized Type I and Type II error rates from Table~\ref{tab:rq2m}.}
\small
\centering
\setlength\tabcolsep{5pt}
\label{tab:error}
\begin{tabular}{l|cc|}
 Treatment & Type I error rate             &Type II error rate  \\\hline
DP unbiased bridge & 0.155 & 0.090 \\
DP biased bridge & 0.095 & 0.075 \\
RSP & 0 & 0.640 \\\hline
\end{tabular}
\end{table}

\noindent\textbf{Answer To RQ2.} In the evaluation of relative bias between decisions made on different data, (1) biased bridge works better than unbiased bridge; (2) relative statistical parity achieves almost identical results as its evaluation between decisions made on the same data; (3) biased bridge has higher power (lower Type II error rate) than relative statistical parity. In conclusion, the biased bridge algorithm performed the best.

\begin{table*}[!tbh]
\caption{Evaluations of differential parity over sensitive attribute $A$ between different decisions made on 2022 Ph.D. applications are shown with numbers of (p value) and d value. Detected differential parity violations are colored as \colorbox{green!20}{green} if $R_i$ prefers $A=1$ more than $R_j$ ($p<0.05$ and $d\ge 0.2$) or colored as \colorbox{red!20}{red} if $R_i$ prefers $A=0$ more than $R_j$ ($p<0.05$ and $d\le 0.2$).}
\centering
\small
\setlength\tabcolsep{5pt}
\label{tab:rq31}
\begin{tabular}{l|cccccc|}
\diagbox{$R_i\quad$}{$R_j$}           & $R_{\text{Committee}}$ & $R_1$       & $R_2$       & $R_3$       & $R_4$      & $R_5$       \\\hline
$R_{\text{Committee}}$ & (1.00) 0.00    & (0.30) -0.17 & (0.27) 0.19  & (0.10) 0.27  & \cellcolor{green!20} (0.00) 0.52 & (0.12) 0.26  \\
$R_1$         & (0.30) 0.17    & (1.00) 0.00  & \cellcolor{green!20} (0.04) 0.33  & \cellcolor{green!20} (0.01) 0.40  & \cellcolor{green!20} (0.00) 0.56 & (0.01) \cellcolor{green!20} 0.41  \\
$R_2$         & (0.27) -0.19   & \cellcolor{red!20} (0.04) -0.33 & (1.00) 0.00  & (0.46) 0.12  & (0.07) 0.31 & (0.56) 0.10  \\
$R_3$         & (0.10) -0.27   & \cellcolor{red!20} (0.01) -0.40 & (0.46) -0.12 & (1.00) 0.00  & (0.33) 0.16 & (0.77) -0.05 \\
$R_4$         & \cellcolor{red!20} (0.00) -0.52   & \cellcolor{red!20} (0.00) -0.56 & (0.07) -0.31 & (0.33) -0.16 & (1.00) 0.00 & (0.12) -0.26 \\
$R_5$         & (0.12) -0.26   & \cellcolor{red!20} (0.01) -0.41 & (0.56) -0.10 & (0.77) 0.05  & (0.12) 0.26 & (1.00) 0.00 \\\hline
\end{tabular}
\end{table*}

\begin{table*}[!tbh]
\caption{Evaluations of differential parity over sensitive attribute $A$ between different decisions made on 2023 Ph.D. applications are shown with numbers of (p value) and d value. Detected differential parity violations are colored as \colorbox{green!20}{green} if $R_i$ prefers $A=1$ more than $R_j$ ($p<0.05$ and $d\ge 0.2$) or colored as \colorbox{red!20}{red} if $R_i$ prefers $A=0$ more than $R_j$ ($p<0.05$ and $d\le 0.2$).}
\centering
\small
\setlength\tabcolsep{5pt}
\label{tab:rq32}
\begin{tabular}{l|cccccc|}
\diagbox{$R_i\quad$}{$R_j$}                          & $R_{\text{Committee}}$ & $R_2$       & $R_3$       & $R_4$      & $R_5$       & $R_6$       \\\hline
$R_{\text{Committee}}$ & (1.00) 0.00    & (0.86) 0.03  & (0.17) 0.23  & \cellcolor{green!20} (0.00) 0.45 & (0.20) -0.20 & (0.17) -0.22 \\
$R_2$         & (0.86) -0.03   & (1.00) 0.00  & (0.12) 0.26  & \cellcolor{green!20} (0.00) 0.48 & \cellcolor{red!20} (0.05) -0.31 & (0.15) -0.22 \\
$R_3$         & (0.17) -0.23   & (0.12) -0.26 & (1.00) 0.00  & (0.19) 0.21 & \cellcolor{red!20} (0.00) -0.48 & \cellcolor{red!20} (0.02) -0.39 \\
$R_4$         & \cellcolor{red!20} (0.00) -0.45   & \cellcolor{red!20} (0.00) -0.48 & (0.19) -0.21 & (1.00) 0.00 & \cellcolor{red!20} (0.00) -0.68 & \cellcolor{red!20} (0.00) -0.57 \\
$R_5$         & (0.20) 0.20    & \cellcolor{green!20} (0.05) 0.31  & \cellcolor{green!20} (0.00) 0.48  & \cellcolor{green!20} (0.00) 0.68 & (1.00) 0.00  & (0.94) 0.01  \\
$R_6$         & (0.17) 0.22    & (0.15) 0.22  & \cellcolor{green!20} (0.02) 0.39  & \cellcolor{green!20} (0.00) 0.57 & (0.94) -0.01 & (1.00) 0.00 \\\hline
\end{tabular}
\end{table*}

\begin{table*}[!tbh]
\caption{Using the biased bridge algorithm, evaluations of differential parity over sensitive attribute $A$ between $R_i$ (different decisions made on 2022 Ph.D. applications) and $R_j$ (different decisions made on 2023 Ph.D. applications) are shown with numbers of (p value) and d value. Detected differential parity violations are colored as \colorbox{green!20}{green} if $R_i$ prefers $A=1$ more than $R_j$ ($p<0.05$ and $d\ge 0.2$) or colored as \colorbox{red!20}{red} if $R_i$ prefers $A=0$ more than $R_j$ ($p<0.05$ and $d\le 0.2$).}
\centering
\small
\setlength\tabcolsep{5pt}
\label{tab:rq33}
\begin{tabular}{l|ccccc|c|}
\diagbox{$R_i@2022$}{$R_j@2023$}                          & $R_{\text{Committee}}$ & $R_2$       & $R_3$       & $R_4$       & $R_5$       & $R_6$       \\\hline
$R_{\text{Committee}}$ & (0.49) -0.00   & (0.48) 0.01  & (0.18) 0.15  & \cellcolor{green!20} (0.05) 0.26  & (0.22) -0.13 & (0.25) -0.11 \\
$R_2$         & (0.16) -0.17   & (0.17) -0.16 & (0.46) -0.02 & (0.25) 0.11  & \cellcolor{red!20} (0.03) -0.32 & \cellcolor{red!20} (0.05) -0.28 \\
$R_3$         & \cellcolor{red!20} (0.02) -0.34   & \cellcolor{red!20} (0.02) -0.34 & (0.11) -0.21 & (0.36) -0.06 & \cellcolor{red!20} (0.00) -0.52 & \cellcolor{red!20} (0.00) -0.46 \\
$R_4$         & \cellcolor{red!20} (0.01) -0.37   & \cellcolor{red!20} (0.01) -0.38 & (0.08) -0.24 & (0.27) -0.10 & \cellcolor{red!20} (0.00) -0.56 & \cellcolor{red!20} (0.00) -0.48 \\
$R_5$         & (0.10) -0.22   & (0.11) -0.21 & (0.34) -0.07 & (0.35) 0.07  & \cellcolor{red!20} (0.01) -0.38 & \cellcolor{red!20} (0.02) -0.33 \\\hline
$R_1$         & (0.48) 0.01    & (0.45) 0.02  & (0.17) 0.17  & \cellcolor{green!20} (0.04) 0.28  & (0.24) -0.12 & (0.27) -0.10 \\\hline
\end{tabular}
\end{table*}

\noindent \textbf{RQ3: How to use differential parity to analyze relative biases in Ph.D. admission?} 

Table~\ref{tab:rq31} shows the differential parity results between each ratings in 2022. Two significant findings can be derived from Table~\ref{tab:rq31}:
\bi
\item
$R_4$ significantly under-rated applicants with a US degree compared to the committee's consensus decision $R_{\text{committee}}$. Actions could be taken to warn $R_4$ in the future ratings to reduce the bias.
\item
$R_1$ significantly over-rated applicants with a US degree compared to all other raters. However, no significant relative bias is detected between $R_1$ and $R_{\text{committee}}$. This may suggest that, although $R_1$'s decision is different from the other committee members, $R_1$ successfully convinced them to agree with some of the decisions in the discussion of $R_{\text{committee}}$.
\ei

Table~\ref{tab:rq32} shows the differential parity results between each ratings in 2023. From Table~\ref{tab:rq32} we can see:
\bi
\item
$R_4$ still significantly under-rated applicants with a US degree compared to the committee's consensus decision $R_{\text{committee}}$. If relative bias is studied in 2022, this could be avoided by warning $R_4$.
\item
There are relative biased detected between $R_2$, $R_3$, $R_5$, and $R_6$. But through discussion of $R_{\text{committee}}$, they have settled on a high-quality consensus decision. This is validated by the result between $R_{\text{committee}}$ in 2022 and 2023 in Table~\ref{tab:rq33}.
\ei

Table~\ref{tab:rq33} shows the differential parity results between 2022 and 2023. From Table~\ref{tab:rq33} we can see:
\bi
\item
$R_{\text{committee}}$ in 2022 is relatively fair to $R_{\text{committee}}$ in 2023. This confirms that high-quality of $R_{\text{committee}}$ in both years. Thus it is reasonable to treat $R_{\text{committee}}$ as the ground truth.
\item
Relative bias between $R_{\text{committee}}$ in 2022 and $R_4$ in 2023, as well as that $R_{\text{committee}}$ in 2023 and $R_4$ in 2022 are confirmed. These results are consistent with the results in Table~\ref{tab:rq31} and Table~\ref{tab:rq32}. Therefore, biased bridge algorithm is reliable in detecting violations of differential parity between decisions made on different data. As a result, before the discussion of $R_{\text{committee}}$ in 2023, biases between $R_{\text{committee}}$ in 2022 and $R_4$ in 2023 could be analyzed and raise an alarm for $R_4$ to re-consider the ratings to alleviate the bias.
\item
$R_5$ in 2023 increased the ratings on applicants with US degrees compared to its own ratings in 2022. This is the only rater who changed the rating strategy. As a result, $R_5$ in 2023 over-rated applicants with US degrees compared to other raters in 2023 when no relative biases between $R_5$ and other raters were detected in 2022.
\item
Although $R_1$ and $R_6$ have not served in the same year. They are relatively fair to each other according to the biased bridge algorithm.
\ei

\noindent\textbf{Answer To RQ3.} Differential parity analysis can be very useful in real world scenarios such as Ph.D. admission. In this case study, with differential parity analysis, we can learn important information such as (1) The final ratings from the committee $R_{\text{committee}}$ are reliable; (2) $R_4$ is relatively biased compared to $R_{\text{committee}}$ and this could be avoided by analyzing the differential parity between $R_4$ and $R_{\text{committee}}$ in 2022 or by analyzing the differential parity between $R_4$ in 2023 and $R_{\text{committee}}$ in 2022; (3) From 2022 to 2023, $R_5$ has changed the way he or she rated the applicants, which should be avoided.

\section{Discussion}
\label{sec:discussion}

This section discusses threats to validity and limitations of this work.

\subsection{Threats to Validity}

\noindent \textbf{Conclusion validity: }The treatments (biased bridge and unbiased bridge) have a significant effect on the outcome based on our paired t test shown in RQ2.

\noindent \textbf{Construct validity: }The theoretical analysis in Section~\ref{sect:methodology} ensures that the outcome (differential parity metrics) corresponds to the effect (whether two decision sets are relatively biased). Theorem~\ref{theorem2} and its proof also ensure that the experimental outcome does correspond to the biased bridge algorithm.

\noindent \textbf{Internal validity: }Sampling bias can have an impact on the outcome of our experiments, especially the training and testing split in the face beauty rating case study. We tried our best to reduce this sampling bias by splitting the data randomly. But still, the outcome can be different with different random seeds.

\noindent \textbf{External validity: }This work suffers from external validity, since it only used two case studies. Whether the conclusions can be generalized to other case studies requires further investigation.

\subsection{Limitations}

This work suffers from the following limitations:
\begin{enumerate}[label=\textbf{L\arabic*}]
\item Starting from Definition~\ref{dp_binary}, the metrics for differential parity and the two estimation approaches are currently defined for binary sensitive attributes. New definitions are required for continuous sensitive attributes. We plan to explore such definitions for continuous sensitive attributes in our future work.
\item Theorem~\ref{theorem1} relies on the assumption that the decision differences are i.i.d. and there is a large enough number of samples to evaluate (based on the central limit theorem and the law of large numbers). This assumption is usually satisfied given the independence of the decision-making process. The i.i.d. assumption of prediction errors in Theorem~\ref{theorem2} is also commonly seen in regression analysis.
\item There is a risk of companies/humans using the differential parity of their decisions against one specific reference set to justify their fairness in the decision-making process. Without the definition of fairness for the context and external checking of the reference set, differential parity can be misleading -- a decision set can be biased when it is relatively fair with respect to another biased reference decision set.
\end{enumerate}

\section{Conclusion and Future Work}
\label{sec:Conclusions}
In summary, this paper proposes differential parity $(R_0-R_1)\perp A$ and relative statistical parity $(\overline{R}_0-\overline{R}_1)\perp A$ to evaluate the relative bias between two sets of decisions -- whether one decision set $R_0$ is more biased towards a certain sensitive group $A$ than another decision set $R_1$. These relative fairness notions alleviate the need to strictly define what is considered to be absolutely fair. In addition, if there exists a reference set of decisions that is well accepted to be fair in a specific context, other decisions satisfying differential parity and relative statistical parity against the reference set can also be considered fair. Next, two novel machine learning-based approaches are proposed to enable the evaluation of differential parity between two decision sets made on different data. This extends the application scenarios of differential parity as shown in Table~\ref{tab:scenario}. In the end, empirical experiments demonstrated the robustness of differential parity metrics, the effectiveness of the machine learning-based approaches, and that differential parity has higher statistical power than relative statistical parity. Another real world case study on Ph.D. admission further illustrates how differential parity can be applied to analyze relative biases between different human raters within the same year as well as across different years.

To resolve the limitations and validity threats in Section~\ref{sec:discussion}, we will explore the future work:
\bi
\item
The metrics of differential parity and relative statistical parity for continuous sensitive attributes.
\item
Applications of the proposed relative fairness testing framework in other realistic scenarios where humans and AI collaborate in decision making.
\item
Conduct human subjects experiments to evaluate the usefulness of differential parity in real time applications.
\ei
Overall, we believe that this work will benefit the artificial intelligence research community by presenting a new way of analyzing relative biases in decisions.

\begin{acks}
This work is funded by NSF grant 2245796.
\end{acks}

\printbibliography

\end{document}